\documentclass[11pt,a4paper]{article}
\pdfoutput=1

\usepackage[hyperref]{acl2022}
\usepackage{times}
\usepackage{latexsym}

\usepackage{microtype}
\usepackage{amsmath}
\usepackage{cleveref}
\crefname{section}{§}{§§}

\aclfinalcopy 

\usepackage{adjustbox}
\usepackage{booktabs}
\usepackage{multirow}
\usepackage{microtype}
\usepackage{xspace}
\usepackage{mathtools}
\usepackage{amssymb}
\usepackage{pifont}%
\usepackage{footnote}
\usepackage{tablefootnote}
\usepackage{graphicx}
\usepackage{color, colortbl}
\usepackage{xstring}
\usepackage{comment}
\usepackage{adjustbox}
\usepackage{float}
\restylefloat{table}
\usepackage{array,booktabs,makecell}
\usepackage{appendix}
\usepackage[normalem]{ulem}
\usepackage[show]{boxnotes}
\usepackage{tcolorbox}
\usepackage{booktabs}
\usepackage{linguex}
\usepackage[utf8]{inputenc}
\usepackage[margin=1.2in]{geometry}
\usepackage[T1]{fontenc}
\usepackage{expex}
\usepackage{tabto}
\usepackage{gb4e}
\usepackage{multicol}
\noautomath
\usepackage{enumitem}
\usepackage{epstopdf}
\usepackage{natbib}
\usepackage{tipa}
\usepackage[british]{babel}
\usepackage{hhline}
\usepackage{qtree} 
\usepackage{tikz}
\usepackage{longtable}
\usepackage{xcolor}
\usepackage{silence}
\def\checkmark{\tikz\fill[scale=0.4](0,.35) -- (.25,0) -- (1,.7) -- (.25,.15) -- cycle;}

\usepackage{scalerel,xparse}

\newcolumntype{H}{>{\setbox0=\hbox\bgroup}c<{\egroup}@{}}

\usepackage{arydshln}

\PassOptionsToPackage{hyphens}{url}\usepackage{hyperref}
  
\title{Towards Afrocentric NLP for African Languages: \\Where We Are and Where We Can Go}

\author{Ife Adebara \\
  Deep Learning and \\Natural Language Processing Group \\
  The University of British Columbia \\

  \texttt{ife.adebara@ubc.ca} \\\And
  Muhammad Abdul-Mageed \\
  Deep Learning and \\Natural Language Processing Group \\
  The University of British Columbia \\

  \texttt{muhammad.mageed@ubc.ca} \\}

\begin{document}

\maketitle

\begin{abstract}

Aligning with ACL 2022 special Theme on ``Language Diversity: from Low Resource to Endangered Languages", we discuss the major linguistic and sociopolitical challenges facing development of NLP technologies for African languages. Situating African languages in a typological framework, we discuss how the particulars of these languages can be harnessed. To facilitate future research, we also highlight current efforts, communities, venues, datasets, and tools. Our main objective is to motivate and advocate for an Afrocentric approach to technology development. With this in mind, we recommend \textit{what} technologies to build and \textit{how} to build, evaluate, and deploy them based on the needs of local African communities.
\end{abstract}

	
\section{Introduction}\label{sec:intro}

Language is the foundation on which communication rests, allowing us to share ideas and interact with one another. Cultures are built on this foundation. We cannot understand, nurture, or help a culture thrive without understanding and nurturing the language carrying it. Language, in turn, is incubated and evolved by culture~\cite{fourie1995introduction}. 
Each culture is thus naturally best expressed using the language in which it evolved, which encodes knowledge about people, their traditions, wisdom, environment, and how they interact with the sum of the concepts that belong to their own culture. Technology is an element of culture that arguably both \textit{shapes} and \textit{is shaped by} it. Technology interacts in complex ways with other elements of culture such as gender, race, and class. Natural language processing (NLP) technologies are no exception, and play an increasingly important role in today's world. 
\begin{figure}{}
\centering
\begin{adjustbox}{width=7cm}
\renewcommand{\arraystretch}{1.2}
{
  \includegraphics[width=10cm,height=8cm]{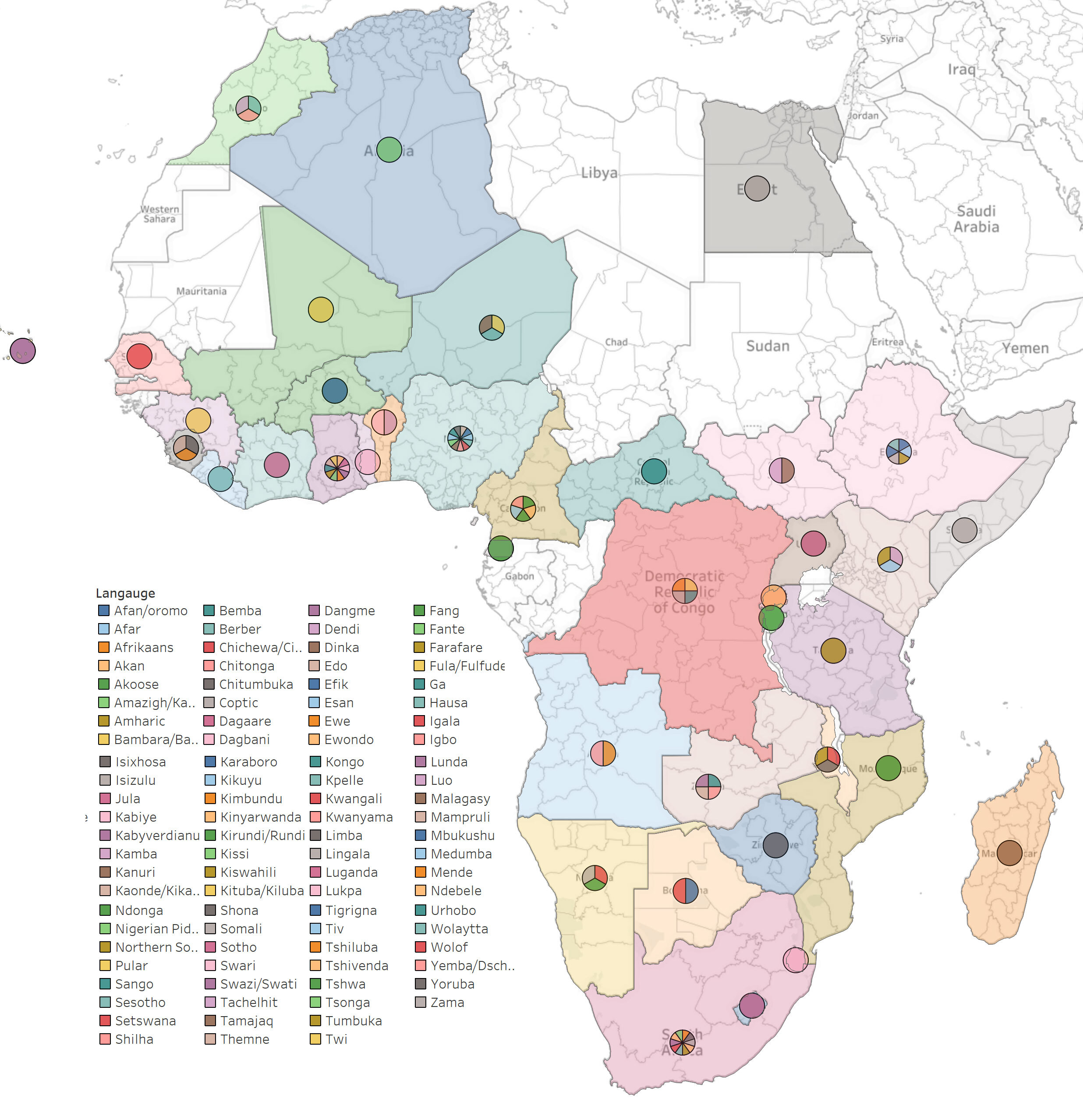}}
  \end{adjustbox}
  \caption{African languages discussed in this paper. A high quality version is in Figure \ref{fig:quality_our_map} (Appendix).}
\label{fig:our_map}
\end{figure}
Modern NLP technologies, however, have primarily been developed in Western societies. As such, they often function within contexts of values, norms, and beliefs that reflect these societies and serve their needs. On the other hand, the very methods employed to develop most of these technologies and the knowledge on which they rest also derive from the same Western-Centric approaches. This poses challenges to the extension and use of these technologies in communities with different social fabrics that speak different languages. The scale of this problem is huge, because the majority of the world's 7000+ living languages~\cite{ethnologue} are not NLP-supported. Apart from perhaps two dozens of popular languages, most languages of the world are under-resourced, indigenous, and/or endangered. Most African languages fall within this category and are the focus of this paper (Figure~\ref{fig:our_map}).

Our goal is to discuss the major linguistic and sociopolitical challenges facing development of NLP technologies for African languages.\footnote{We do not cover Arabic since it is spread in both Africa and Asia, and has a sizeable NLP community.} In doing so, we both motivate and advocate for an Afrocentric approach to technology development where \textit{what} technologies to build and \textit{how} to build, evaluate, and deploy them arise from the needs of local African communities. We start by typologically situating African languages and providing illustrating examples as to what makes them challenging from a computational linguistics perspective~(\cref{sec:afri_specs}). Next, we discuss consequences of the literacy situation in Africa on NLP~(\cref{sec:disruptive_tech}). We then further explain why the classical binary approach to technology development of \textit{feature engineering} vs. \textit{end-to-end} solutions familiar to most NLP researchers is not ideal for the African context~(\cref{sec:feng_end2end}). We follow by data quality~(\cref{sec:data-quality}). To facilitate future work, we also point to ongoing community efforts, venues, and datasets~(\cref{sec:community}). 
We conclude in~\cref{sec:conc}. 
\section{Why Typology Matters}\label{sec:afri_specs}


Although it has been argued that the best way to achieve cross-linguistically useful NLP is to leverage findings of typological research~\cite{bender2016linguistic}, most NLP work remains Indo-Eurocentric in terms of algorithms for preprocessing, training, and evaluation. This is a mismatch to the fact that every NLP approach requires either explicit or implicit representative linguistic knowledge~\cite{o2016survey,ponti2019modeling,bender2016linguistic}. Knowledge of linguistic typology can indeed be very useful for both language-specific and language-independent NLP~\cite{o2016survey}, including for African languages. This knowledge can be useful for determining which \textit{languages} may be treated together (e.g., in multilingual models) and/or which \textit{methods} are best suited for a language-specific task (e.g., a method can be deemed potentially useful if it has been applied successfully on a language with a similar typology). To illustrate what typological information can concretely mean for African languages, it may be useful here to list a number of the most notable typological features prevalent in African languages across several language families including \textit{Afro-Asiatic, Austronesian, Niger-Congo, Nilo-Saharan, Indo-European} and \textit{Creole}. These features include use of tone, open syllables, vowel harmony, splitting verbs, serial verb construction, reduplication, use of very few or no adjectives (closed class of adjectives \cite{Guillaume}), and a large number of ideophones. 
We will discuss three of these features which we judge as largely absent from most of the top $10$ NLP-popular languages.\footnote{We use the language diversity index of~\citet{joshi-etal-2020-state} to select the top $10$ languages.} We provide a list indicating presence of one or more of these features in over $100$ African languages in Appendix Table \ref{tab:lang-info}. 


\subsection{Tone} 
Phonemic tone is characteristic of many African languages, with $\sim80\%$ of these languages being tone languages~\cite{hyman2003african, creissels2008africa}. This includes most languages of the Niger-Congo family, except Swahili, Wolof, Serer, Cangin, and Fulani which are not tone languages. All Nilotic and Khoisan languages and many Afroasiatic languages are also tonal. A smaller number of languages, including Somali and many Bantu languages, are tonal accent languages, in which a distinctive or demarcative accent is expressed by a toneme of high pitch \cite{clements_rialland_2007}. 

Tone can occur both at the lexical and grammatical levels. Lexical tones are a difference in pitch that distinguishes one lexeme from another. In Yor\`{u}b\'{a}, for instance, lexical tone is responsible for the differences in meaning in the following: \texttt{igb\'{a}} (\textit{calabash}, \textit{basket}), \texttt{igba} (\textit{200}), \texttt{\`{i}gb\`{a}} (\textit{time}), \texttt{\`{i}gb\'{a}} (\textit{garden egg}), and \texttt{igb\`{a}} (\textit{rope}). Grammatical tone, on the other hand, distinguishes one grammatical category from another. In Akan, a language with both lexical and grammatical tone, grammatical tone distinguishes habitual and stative verbs as in: \texttt{Ama d\'{a} ha} \textit{`Ama sleeps here'} and \texttt{Ama d\`{a} ha} \textit{`Ama is sleeping here'}. Grammatical tone is also used to indicate case in some Bantu languages~\cite{creissels2008africa, konig2008case}, as a definite marker, for inflectional or derivational purposes, or to code spatial relations \cite{creissels2008africa}. 

Two approaches have been adopted in the orthographies of African tone languages: \textit{no tone marking} or \textit{tone marking}. \textbf{No Tone Marking.} Hausa, spoken in Niger and Nigeria, has grammatical tone but adopts a no tone marking approach in its orthography. This results in ambiguities that may not be resolved in context as in\texttt{ j\'{a}\'{a} t\`{a}f\'{i}} `He went', \texttt{j\'{a}\`{a} t\`{a}f\'{i}} `He may go', and \texttt{j\`{a} t\`{a}f\'{i}} `He should go' ~\cite{cahill2019tone}. It is worth mentioning that no tone marking makes little difference in tone languages with few minimal pairs. NLP systems designed for a tone language without tone marking may therefore suffer from issues with ambiguity, if contextual information is not adequate for disambiguation or if many minimal pairs exist in the language.


\textbf{Tone Marking.} Languages that mark tone may adopt a \textit{shallow} marking (Yor\`{u}b\'{a}) or \textit{deep} marking approach~\cite{cahill2019tone, steve_bird_1999_b} by using diacritics, punctuation marks, or letters to indicate tone~\cite{cahill2019tone}. A shallow marking approach uses the surface level tone after phonological rules (such as assimilation) that change the representation of tones have been applied. The implication of this type of approach is that the same word will have different tone representations in different contexts. In a low-resource scenario, therefore, each word will have fewer occurrences and some contexts may not be seen in training data~\cite{steve_bird_1999} (i.e., data sparsity). For languages that adopt a deep marking approach, a word would have the same tone, \textit{orthographically}, in every context. However, the speech token representing the same word will vary, thus creating ambiguity at the speech front. Although adopting a shallow or deep marking approach may not have significant implications on languages with few tone phonological rules, the degree of shallow-to-deep marking may increase ambiguity for languages with many phonological rules~\cite{steve_bird_1999, steve_bird_1999_b}. Tone-marking can also be \textit{partial} or \textit{exhaustive}. \textbf{Partial Tone-marking.} Some African languages such as Yor\`{u}b\'{a} adopt a partial tone marking approach with diacritics. Yor\`{u}b\'{a} has three distinctive tones - high, mid and low tones - but only represents the high tone with the acute symbol and the low tone with the grave symbol in its orthography. The mid tone is not marked and vowels without diacritics unambiguously indicate the presence of the mid tone. Rangi, a language spoken in Tanzania, marks only high tone on nouns while Akoose, a language spoken in Cameroon, marks high tone and contour tones but leaves low tones unmarked \cite{cahill2019tone}. Karaboro, spoken in Burkina faso, marks grammatical tones in plurals using a word final hyphen as in:  \texttt{s\`{a}\`{a}pj\'{e}} `Rabbit' and \texttt{s\`{a}\'{a}pj\'{e}-} `Rabbits'. \textbf{Exhaustive Tone-marking.} In exhaustive tone-marking, every tone bearing unit is orthographically marked for tone as in Dschang, spoken in Cameroon~\cite{steve_bird_1999}.

Furthermore, a higher number of distinctive tones increases ambiguity. In Dan, a language with five distinctive tones, the following can occur: \texttt{gba$^1$} \textit{(caterpillar)}, \texttt{gba$^2$} \textit{(shelter)}, \texttt{gba$^3$} \textit{(fine)}, \texttt{gba$^4$} \textit{(roof)}, and \texttt{gba$^5$} \textit{(antelope)} \cite{clements2008africa}. For another example, Yor\`{u}b\'{a} has three distinctive tones where each monosyllabic sequence of sounds can have up to three pitch contrasts and a bi-syllabic can have $2^3$ pitch contrasts.\footnote{However, some phonological rules can restrict the occurrence of certain combinations and there may be lexical gaps. For instance, the high tone occurs only in marked consonant-initial words.}

\textbf{Recommendations.}
\textbf{(1)} For speech applications, there exists a plethora of unexplored research questions to answer with regard to the implication of tone on \textit{text-to-speech} and \textit{text-free speech processing}~\cite{lakhotia2021generative}. We therefore call for empirical studies that investigate the influence of tones in text-to-speech, text-free speech processing, and universal speech processing~\cite{yang2021superb}. Since tone is absent in Indo-European languages where most recent speech work is situated, we expect this to be a fruitful direction. 
\textbf{(2)} For text applications, tone will be relevant for \textit{natural language understanding (NLU)} tasks including but not limited to \textit{part of speech tagging (POS)}, \textit{text classification}, and \textit{natural language generation (NLG)} tasks such as \textit{machine translation}. For many of these applications, it is not clear how tone would interact with system performance. For example, we do not know where to keep and where to remove tone (if at all). For example, we find that while removing tone has negligible impact on Bambara$\rightarrow$English MT, it has significant negative impact on Y{o}r\`{u}b\'{a}$\rightarrow$English (see Table~\ref{tab:bleu} in Appendix). We also do not necessarily know what the best ways to encode (and decode) tone information are.  
\textbf{(3)} For work involving languages with shallow tone marking at the orthographic level, we recommend budgeting for collection and preparation (e.g., annotation) for sizeable datasets (to alleviate data sparsity). In absence of large datasets, knowledge of the finite phonological rules of a language can also be exploited for generating data for downstream tasks. 
\textbf{(4)} Orthographic conventions should not be taken as a good indication of the functional load (i.e., information load) of tone in a language, for there are many non-linguistic (e.g., political) reasons for employing a particular orthographic convention~\cite{m_cahill_2011}. Hence, NLP researchers should do due diligence as to understanding how tone works in a given language. 
\textbf{(5)} Punctuation marks may be tone indicators, and care needs to be taken on how these are pre-processed. 

\subsection{Vowel Harmony} 
Vowel harmony is a phonological pattern in which vowels within a given domain agree in properties such as tongue position or lip rounding \cite{hyman2003african}. It restricts the possibilities of vowels that can co-occur \cite{archangeli2007harmony}. Different languages adopt different types of vowel harmony. Three types of vowel harmony that are unique to African languages have been recorded in the literature~\cite{clements_rialland_2007}: (i) advanced tongue root (ATR) harmony, (ii) cross height ATR harmony, and (iii) reduced ATR harmony. ATR harmony occurs when some vowels have the $[-ATR]$ feature and others have the $[+ATR]$ feature. Within a word, all non-low vowels agree in $[+ATR]$ or $[-ATR]$ features. With cross height ATR, $[+ATR]$ in mid vowels require  $[+ATR]$ in high vowels and vice versa. The reduced ATR, on the other hand, occurs in languages with only one mid vowel and $[-ATR]$ mid and high vowels shift to $[+ATR]$ in the context of $[+ATR]$ high vowels \cite{clements_rialland_2007}. 

\textbf{Recommendations.} \textbf{(1)} Since vowel harmony is largely absent in most Indo-European languages, knowledge of vowel harmony is currently underexplored in NLP. Such a knowledge can be useful for tasks such as \textit{POS tagging} since tokens with the same part of speech tend to have similar harmonies. \textbf{(2)} \textit{Automatic spelling checkers} can also exploit information about vowel harmony since certain co-occurrences of vowels are barred by phonological rules of vowel harmony.

\subsection{Serial Verb Constructions} 
Serial verb constructions (SVC) involve two or more verbs that combine as a whole without any indication of dependency or any conjunction between them \cite{creissels2008africa, dechaine2008serial}. Languages with SVC use serial verbs to encode events that are usually encoded as single verbs in Indo-European languages. This poses a unique problem when creating/evaluating \textit{cross-lingual embeddings} and in applications such as \textit{dictionary creation}. For instance translating from English to Yor\`{u}b\'{a}, we have the following examples: \textit{borrow} - `\texttt{Gb\`{a} \`{a}w\`{i}n} (receive credit)', \textit{believe} - `\texttt{Gb\`{a} gb\textsubdot{\'{o}}} (receive hear)', \textit{pinch} - `\texttt{J\'{a} l' \'{e}\`{e}\'{e}k\'{a}n\'{a}} (cut with fingernails)' so that a single English verb is a serial verb in Yor\`{u}b\'{a}. When these words are used in sentences, they may have intervening words as in: \texttt{Gb\`{a} \'{a} l' \'{a}aw\`{i}n} (receive 3SG-O on credit) \textit{`borrow it'}, \texttt{Gb\`{a} \'{a} gb\textsubdot{\'{o}}} (receive 3SG-O hear) \textit{`believe it'},  \texttt{J\'{a} a  l' \'{e}\`{e}\'{e}k\'{a}n\'{a}} (cut 3SG-O with fingernails) \textit{`pinch it'}. In Africa, serial verb constructions are very common in Kwa (e.g. Ewe) and Western Benue-Congo languages (e.g. Yor\`{u}b\'{a}). They have also been recognized in the North Khoisan language !Xun. 

\textbf{Recommendations.} \textbf{(1)} Given how pervasive word embedding models are in most NLP applications, we recommend investigating how embeddings accounting for SVC can be developed. Similarly, SVC will have bearings in how (cross-lingual) embeddings are evaluated. For example, researchers may need to create dictionaries customized to African languages. \textbf{(2)} For \textit{POS tagging}, decisions need to be made on what approach to take in treating such constructions. \textbf{(3)} Research investigating the extent to which SVC affects performance across different tasks needs to be explored. For example, this can be valuable for \textit{parsing} and \textit{MT}. 




\section{No Literacy, No NLP}\label{sec:disruptive_tech}
NLP for high resource languages (HRL) benefits from the level of literacy NLP researchers have in these languages. Most researchers usually have literacy beyond high school in one or more of the languages they work on. In Africa, however, with very complex multilingual societies, many educated Africans cannot read nor write their Indigenous languages.\footnote{We use \textit{Indigenous} languages to refer to languages native to Africa.} These people do not have basic linguistic knowledge in their languages either. For example, many people do not know which words are nouns or verbs~\cite{cahill2001avoiding}.
For context, more than $2,000$ languages have been reported in Africa - about $1/3$ of all the languages in the world~\cite{hammarstrom20181} - making many African communities truly multilingual. As a result, it is not uncommon for a child to be exposed to multiple Indigenous languages before reaching school age. This is especially the case in families where the father, mother, and grandparents all speak different languages (which may, in turn, be different from the languages spoken in the communities they live in). People who receive formal education - the sole way people become literate - thus attain only partial literacy in one or more African language(s) which may not even be their mother tongues. Many others have no knowledge of any Indigenous language, and are only literate in a foreign language~\cite{cahill2001avoiding, ouane2010and}.

As seen in Table \ref{tab:lang-use} in the Appendix, out of the $56$ countries in Africa, only $17$ countries have an Indigenous language as a national language (although in $14$ of these $17$ countries, a foreign language is the main official language). Furthermore, the countries that give any official status to Indigenous languages, tend to restrict such a status to those languages belonging to majority speakers.\footnote{It is worth mentioning that some of the excluded languages have millions of L1 speakers.} For example, in Nigeria, only three out of $512$ languages are officially recognized as regional languages; Ghana uses $10$ of its $73$ Indigenous languages as institutional languages; Swahili is the only official Indigenous language in Tanzania out of $118$ others; $12$ of $61$ languages in Kenya have some official status; only $12$ of $20$ Indigenous languages in South Africa are institutional languages. This challenging situation is the result of poor language policies, which we now turn to.

\textbf{Language policy.} Language policy determines which languages are used in education, media, commerce, and almost every domain controlled by government. With most Africans educated in English, French, Portuguese or majority African languages, most African languages (those without any official status) are rarely used or used only at home~\cite{petzell2012linguistic, foster2021language, ouane2010and}. In countries where an Indigenous language has official status, governments and implementing bodies only pay lip service to these policies~\cite{ThePoliticsofLanguageEducationinAfrica}. In addition, lack of trained personnel and adequate educational resources in Indigenous languages, as well as rarity of teachers sufficiently proficient to offer Indigenous language courses, make policies difficult to implement~\cite{trudell2018language, ThePoliticsofLanguageEducationinAfrica}. Furthermore, in many schools, Indigenous languages are referred to as vernaculars and are prohibited. Violation usually attracts fines, and even corporal punishment in some cases. English and other foreign languages remain the prerequisite for scientific and technological development, and a key to social prestige and power. Students who do not pass examinations in these foreign languages cannot continue studying beyond elementary school~\cite{foster2021language, petzell2012linguistic, mohr2018changing}. Effect of these currently implemented policies is visible in the NLP situation of African languages. Languages officially recognized within their countries have more resources and tools for NLP than those that do not. For instance, all African languages with a diversity index~\cite{joshi-etal-2020-state} greater than zero are either official national, regional, or educational languages or are languages of wider communication~\cite{ethnologue}. We provide more details about available resources of different types (labelled, unlabelled, parallel, and raw) and tools in Section~\ref{subsec:resources} (Appendix).

\textbf{Recommendations.} Partial and lack of literacy or knowledge of Indigenous languages has significant negative impacts on NLP in African languages. Therefore, \textbf{(1)} we include in our concept of a grand challenge the development of language policies that facilitate literacy in Indigenous African languages. Literacy improvement takes time, and policies that teach Indigenous languages only for brief periods in elementary school need to be reformed. \textbf{(2)} We also recommend the implementation of policies that require use of Indigenous languages in media, government, and other domains. \textbf{(3)} Adequate funding needs to be allocated to develop pedagogical materials, train teachers, and provide teaching aids in order to facilitate the implementation of these policies. \textit{Simply put, without improvement of literacy in African languages, we do not see a flourishing future for African NLP.}

\section{A Tale of Two Approaches}\label{sec:feng_end2end} 


There are two main approaches for developing NLP systems. We discuss each of these \textit{vis-a-vis} the situation for African languages here, giving relevant recommendations.

\noindent\textbf{Feature engineering}. Feature engineering requires domain knowledge, which is lacking for many African languages due to the aforementioned literacy situation. This negatively impacts use of written African languages in many domains of human endeavor, let alone NLP research. Weak literacy simply means unavailability and inaccessibility of linguists, annotators, language experts, and computational linguists with expertise in African languages. It also manifests itself in lack of grammars, primers, teaching aids, and dictionaries~\cite{m_cahill_2011}. As it turns out, grammatical information is either fully lacking or under-documented for almost half of Africa's languages. This makes Africa the second \textit{least known} continent (after Oceania, dominated by the New Guinea area)~\cite{hock2011world}. In Appendix Table~\ref{tab:lang_resources}, we list available linguistic resources for all African languages we could trace. 

\noindent\textbf{Deep Learning Approaches.} A major bottleneck in the development of end-to-end deep learning NLP systems for African languages is the paucity of machine-readable data~\cite{adda2016breaking}. Deep learning systems for high-resource languages are usually fed ever-growing amounts of data that are abundant online and via several other avenues in today's connected society. Without these type of (interactive) data, it is challenging to develop NLP models for real-world use. In particular, models that are endowed with the implicit and explicit knowledge embedded in language are hard to build (at least by current technologies) without large volumes of data derived from diverse contexts.
Many African languages lack the environment from which these types of machine-readable data can be collected. Social media, which is a venue for data collection for many high-resource languages, are often not widely used for African languages. In fact, most Africans post to social media in foreign languages rather than in Indigenous African languages~\cite{malatji_2019}.\footnote{https://www.talkwalker.com/quick-search.} One reason behind this issue is unavailability of keyboards for Indigenous languages. Most keyboards, for example, do not support symbols for representing tone and some other grammatical features. Partial or complete lack of writing literacy is another reason. A third reason is related to the lack access to smart machines and internet connectivity.

Furthermore, countries such as Nigeria where official status is given to a handful of Indigenous languages, still document official activities in foreign language exclusively. Media organizations that often read the news in a foreign language as well as local languages also archive \textit{only} the English news and discard those in Indigenous languages. All such practices stifle opportunities for developing large datasets for African languages, effectively causing African NLP to lag behind. If archived, data for many Indigenous African languages can facilitate development across a wide host of speech and language tasks, including text-to-speech and machine translation. Collectively, these compounded issues mean there are only few (and often smaller) online communities that contribute to web fora, Wikipedias, and other platforms where data are growing in large-to-massive amounts for high-resource languages. This is evident in the diversity index for African languages offered by~\newcite{joshi-etal-2020-state}. 

According to~\newcite{joshi-etal-2020-state} who summarized the digital status and `richness' of languages in the context of data availability, $542$ African languages are \texttt{left-behinds}. That is, these languages have exceptionally limited resources that will make it probably impossible to lift them up in the digital space. A total of $26$ African languages are \texttt{scraping-bys} and are in a better position than the \texttt{left-behinds}. However, even these are said to require organized awareness and strong data collection effort with most of these languages having no labelled datasets. Only nine African languages are in the \texttt{hopefuls} category, with a small set of labeled datasets, researchers, and language support communities. A single African language (i.e., Afrikaans) is in the \texttt{rising-stars} category with a strong web presence and a thriving cultural community online (although with insufficient efforts in labeled data collection). We offer a summary of the diversity index for $578$ African languages in Table~\ref{tab:l_index} in the Appendix. 

\noindent\textbf{Recommendations.} \textbf{(1)} We recommend that daily engagements in education, commerce, media, and government which are otherwise archived only in foreign languages (see Table \ref{tab:lang-use}), be archived in Indigenous languages as well. These would comprise valuable sources of labelled and unlabelled machine-readable data for NLP, let alone painting a more equitable and representative picture of African languages. \textbf{(2)} Humans and machines complement each other’s strengths, so we recommend stronger interactions between NLP experts and theoretical linguists or knowledgeable native speakers when developing resources and models for African languages. \textbf{(3)} Funding should also be allocated to theoretical linguists and language experts, along with machine learning and NLP experts, to aid this work.  \textbf{(3)} For African languages with available linguistic research, it has been found that certain POS, morphological, named entity, and dependency information can be accurately retrieved \textit{automatically} by using tone, vowel harmony, or even syllable structure patterns~\cite{10.1145/2872518.2890563}. These approaches may aid faster development of POS taggers, lemmatizers, NER, or even dependency parsers. \textbf{(4)} When developing NLP pipelines for African languages, removal of numbers and non-alphanumeric symbols should be approached with caution. This should especially be the case for languages with insufficient research as to the functions played by these symbols, and would help avoid making any irrecoverable issues in the data. \textbf{(5)} The most effective ways for building pipelines for African languages remains an under-explored area of research. We therefore call for empirical studies that investigate development of viable pipelines. \textbf{(6)} We emphasize the need to respect user consent, data sovereignty, wishes of local communities, and other important issues such as privacy while carrying out any collection or archival effort~\cite{sutherland2018digital, daigle2021data, 10.1093/idpl/ips014}. This is to prevent the predatory use of data collected from local communities including monitoring or controlling local peoples, censorship, and other surveillance activities. Properly handling data mitigates physical, financial, and other security risks that poor data practices expose local communities to~\cite{turianskyi2018balancing} and must also be prioritized. We now further discuss issues around data quality.

\section{Garbage in, Garbage out}\label{sec:data-quality}
A manual evaluation of $205$ datasets involving African languages such as those in CCAligned \cite{elkishky_ccaligned_2020}, ParaCrawl \cite{banon-etal-2020-paracrawl, espla2019paracrawl}, WikiMatrix \cite{schwenk2019wikimatrix}, OSCAR \cite{suarez2020monolingual}, and mC4 \cite{xue-etal-2021-mt5} show that at least $15$ corpora were completely erroneous, a significant fraction contained less than $50\%$ of correct data, and $82$ corpora were mislabelled or used ambiguous language codes \cite{caswell2021quality}. The inaccuracy is due to a lack, or poor quality of language identification tools, dictionaries, and text pre-processing piplelines, for many low resource languages including African languages represented in these datasets. Furthermore, available resources are rarely evaluated especially when crawled as part of a multilingual dataset. Furthermore,~\newcite{alabi2020massive} find that, fastText embeddings for Yor\`{u}b\'{a} has an estimated 135K out of 150K words belonging to other languages such as English, French, and Arabic. New embedding models created by ~\newcite{alabi2020massive} with a curated high quality dataset outperform the off-the-shelf fastText embeddings even though the curated set has fewer words. Results of these few studies paint a gloomy picture for most current multilingual datasets involving African languages, and models derived from them. 

Inconsistent orthographies also contribute to the data quality problem \cite{martinus2019focus}. In many cases, orthographies may not be standardized and will have significant spelling and punctuation variations across different domains. In some cases where standard orthographies exist, word lists or dictionaries do not necessarily represent the standardized orthography. Using Hausa as an example, all commercially published books and nearly all Hausa language newspapers use the standard romanized orthography. Standard romanized orthography is written without tones or any indication of vowel length~\cite{schuh_yalwa_1993}. The orthography used in grammars, dictionaries, and pedagogical documents on the other hand, indicate tone and vowel length~\cite{schuh_yalwa_1993}. Furthermore, languages that have standard orthographies may also suffer from inconsistencies when orthographic conventions are not adhered to~\cite{olumuyiwa2013yoruba}. This is evident in the methods and practices for content archiving of many African languages on the web. For example, all VOA websites, omit tones for African languages whose standard orthographies require tone diacritics. BBC also does not adhere to the orthographic conventions for Yor\`{u}b\'{a} texts except in the headlines, JW.org also does the same for some African languages. 

Apart from the aforementioned issues, lack of constant and systematic use of African languages in contexts such as governance, law, technology, science, and education prevents African languages from expanding in vocabulary to accommodate new concepts that have become important parts of conversation elsewhere. As a result, it is not uncommon to have large amounts of foreign words in a dataset which are not adapted to the phonological or orthographic structure of the target African language. Furthermore, terminologies continue to be employed inconsistently and spelt differently in many African venues.

To provide a concrete example of the data quality problem for African languages, we perform a manual evaluation of Flores-101 dataset~\cite{flores101, flores101b} for Y{o}r\`{u}b\'{a}. We find the following: \textbf{(1)} $5.29\%$ spelling errors \textbf{(2)} $2.7\%$ inconsistent spellings \textbf{(3)} $1.2\%$ borrowed words not adapted to the orthographic conventions of target language and \textbf{(4)} $12.4\%$ incorrect tone marks. Detailed information is in Appendix~\ref{sec:dataquality}.

It is important to mention that a single error in assignment of diacritics, for instance, can result in significant semantic and syntactic differences in texts. The implication of inconsistencies in orthography is hence enormous for low resource African languages. Such inconsistencies worsen the issue of \textit{data sparsity}: when different spellings of the same word are employed, or when tone or other grammatical features are inconsistently marked, the same `word' will have many more surface forms than what it actually should. Data sparsity can in turn aggravate the situation for any work involving training with data from different domains (e.g., in domain adaptation). That is, reliability of models trained with erroneous data from a source domain will be diminished while transferring into a target domain. Orthographic inconsistencies also affect results of search engines~\cite{choros2005testing} in that these engines would not recognize the relationship between a diacritized text and its undiacritized counterparts~\cite{asubiaro2014effects, olumuyiwa2013yoruba}. Again, this results in difficulty retrieving resources for many African languages. To optimize search for African languages that involve diacritics, some users employ normalized text which in turn further creates a mismatch between web documents and other standard offline documents (e.g., books) for many African languages.

\noindent\textbf{Recommendations.} \textbf{(1)} We recommend developing language identification tools that cover African languages. \textbf{(2)} Development of dictionaries or even extended word lists will also help the community ensure data quality. \textbf{(3)} Manual inspection of sizeable samples of multilingual datasets should also continue to be prioritized. \textbf{(4)} We also suggest orchestrated efforts to enforce consistency in orthography for the various languages. \textbf{(5)} Linguistic rules may be appropriate for developing automatic data cleaning and pre-processing, but development of any such rules should be carried out carefully. We now briefly highlight community efforts invested in developing skills, datasets, and tools in the African NLP space.


\section{Communities and Resources}\label{sec:community}
\textcolor{red}{}The majority of existing resources for NLP are the initiative of various non-governmental organizations determined to develop datasets and tools for African languages. We list some of these efforts for NLP, but also within the larger contexts of artificial intelligence. We focus on communities and venues here and list recent funding initiatives in Table~\ref{tab:funding} (Appendix).

\noindent\textbf{Workshops.}
As far as we know, there are two main venues in the form of workshops supporting NLP for African languages, and African AI. These are AfricanNLP and BlackInAI. We provide details about these venues in Appendix~\ref{subsec:workshops}. 

\noindent\textbf{Communities.}
\href{https://www.masakhane.io/}{Masakhane}, \href{https://blackinai.github.io/#/}{Black in AI}, \href{https://deeplearningindaba.com/about/our-mission/}{Deep Learning Indaba}, \href{https://www.k4all.org/}{Knowledge 4 All Foundation Ltd (K4A)}, \href{https://zindi.africa/}{Zindi} and \href{http://www.alt-i.org/}{ALTI} are some of the active communities for research on NLP for African languages. More information about each of these communities is in Section~\ref{subsec:community}. 

\noindent\textbf{Resources.}
The religious domain is currently the major source of data for a large number of African languages. Top amongst religious resources is the Bible corpus (available in over $1,000$ languages of Africa~\cite{resnik1999bible,mccarthy2020johns}) and the JW300 website (with data for $\sim100$ low-resource African languages). Religious sources are constantly updated with new data from the same languages and new languages are often added, making these sources increasingly useful. One issue of these datasets is that, although they are parallel, they may not be sentence aligned. Regardless, these resources remain significantly inadequate. Most other data available for African languages are raw and unlabelled. Still, these can be useful in many applications (e.g., in training word embeddings or language models, for backtranslation). We provide more details about available resources (labelled, unlabelled, and raw) and tools in Appendix~\ref{subsec:resources}. 

\noindent\textbf{Recommendations.} \textbf{(1)} To achieve Afrocentric NLP, we recommend active interactions between differently existing communities, as well as encouraging new regional and thematically-defined communities. \textbf{(2)} We recommend extending these communities beyond AI, NLP, and machine learning to involve theoretical linguists, anthropologists, sociologists, field workers, and other scholars and practitioners with interest in African languages. \textbf{(3)} We believe ACL and other similar organizations should continue to prioritize work on low-resource languages by securing dedicated tracks in their publication and dissemination venues.

\section{Discussion and Conclusion}\label{sec:conc}
 We discussed major challenges facing development of NLP technologies for African languages. One of the most important recommendations we would like to emphasize is to \textit{prioritize African NLP work based on the needs of African communities}. For example, we believe development for data and tools for improving health and education should be a priority. We also caution against extractive practices, and encourage creation of opportunities, contexts, and venues for work on African languages and advocacy for reclaiming African language policies. In addition, data literacy and issues around data sovereignty and privacy should remain of highest importance. We highlighted various communities and venues here that we think should continue to be supported.

\section*{Acknowledgements}\label{sec:acknow}
We gratefully acknowledge support from the Natural Sciences and ENgineering Research Council of Canada (NSERC; RGPIN-2018-04267), the Social Sciences and Humanities Research COuncil of Canada (SSHRC; 435-2018-0576; 895-2020-1004; 895-2021-1008), Canadian Foundation of Innovation (CFI; 3771), Compute Canada (CC),\footnote{\href{https://www.computecanada.ca}{https://www.computecanada.ca}{computecanada}} and UBC ARC-Sockeye.\footnote{\href{https://arc.ubc.ca/ubc-arc-sockeye}{https://arc.ubc.ca/ubc-arc-sockeye}} Any opinions, conclusions or recommendations expressed in this material are those of the author(s) and do not necessarily reflect the views of NSERC, SSHRC, CFI, CC or UBC ARC-Sockeye. We thank Rose-Marie D\'{e}chaine for helpful discussions.



\normalem

\bibliographystyle{acl_natbib}
\bibliography{references}
\newpage 
\appendix
\clearpage
\appendixpage
\addappheadtotoc
\numberwithin{figure}{section}
\numberwithin{table}{section}

\section{Effect of Tone in MT}\label{subsec:tone_mt}
In this experiment on tone, we used the bible for the Yor-En pairs \cite{adebaratranslating}, and LDC dataset (Bamanankan Lexicon LDC2016L01.) for the Bam-Eng pairs. Details of the data sizes are available in Table \ref{tab:data}.
\begin{table}[H]
\small 
\centering
\begin{tabular}{l|c|r|r}
 \toprule
\multicolumn{1}{c}{\textbf{Pair}} & \multicolumn{1}{c}{\textbf{Lang}} & \multicolumn{1}{c}{\textbf{Sent}} & \multicolumn{1}{c}{\textbf{Words}} 
\\
 \toprule
\multirow{2}{*} {\textbf{Bam-Eng}}  & \textbf{Bam} & $11,154$ & $43,786$M \\ 
                                  & \textbf{Eng} & $11,154$ & $64,571$ \\
\multirow{2}{*} {\textbf{Yor-Eng}}   & \textbf{Yor}  &$31,086$  & $942,663$\\
                                  & \textbf{Eng}  &$31,086$  &$822,950$   \\
 \toprule
\end{tabular}
\caption{Number of sentences and words for the training data used for each language pair.}\label{tab:data}
\end{table}

We developed python scripts to remove diacritics from Bambara and Yor\`{u}b\'{a} no-tone marked settings. In Table \ref{tab:bleu}, tone significantly affects bleu scores for En-Yor pairs but has marginal effect in the Bam-En pairs. The influence of tones thus needs to be further investigated.
\begin{table}[H]
\small 
\begin{tabular}{>{}c|l|r|rH}
 \toprule
\multicolumn{1}{c}{}  &\textbf{Pair} & \textbf{Tone-Marked} & \textbf{No-Tone Mark}  \\ 
 \midrule

\multicolumn{1}{c}{}  & BAM-ENG & $1.61$ &  $1.61$ \\

\multicolumn{1}{c}{}  &ENG-BAM & $1.07$ &  $1.34$ \\

\multicolumn{1}{c}{}  &ENG-YOR & $32.95$ &  $11.51$ \\

\multicolumn{1}{c}{}  &YOR-ENG & $38.57$ &  $12.76$ \\



 \bottomrule
\end{tabular}
\caption{BLEU scores for tone-marked and no tone mark settings.} \label{tab:bleu}
\end{table}

\section{Language Typology Information}
In Table~\ref{tab:lang-info}, we provide typology information covering tone, vowel harmony and SVC for $116$ African languages. The checkmarks indicate the presence of the specified feature in the language. To the best of our ability, this information represents the features in the specified languages and for the specified features. Although we do not claim that this information is complete. This table was created by perusing grammatical descriptions, pedagogical materials, and linguistic research regarding these features in the specified languages.

 \begin{table*}[h!]
\small
\centering
\begin{adjustbox}{max width=\textwidth}
\renewcommand{\arraystretch}{1.15}
{
        \begin{tabular}{>{}cllllll|llllllll}
        \toprule

      \textbf { }    &\textbf { \small Language }   &\textbf {\small Code} & \textbf{ \small Tone} & \textbf{ \small T.Marked} & \textbf{ \small VH} & \textbf{ \small SVC}  &\textbf { \small Language }   &\textbf {\small Code} & \textbf{ \small Tone} & \textbf{ \small T.Marked} & \textbf{ \small VH} & \textbf{ \small SVC} \\    \midrule
  \multicolumn{1}{c}{}  & \multirow{1}{*}{Afar} & aar & && &&{Amharic} & amh &&&\checkmark \\
 \multicolumn{1}{c}{}  & \multirow{1}{*}{Amazigh} & kab &&&&&{Coptic} & cop & &&  \\
   \multicolumn{1}{c}{}  & \multirow{1}{*}{Ge'ez } & gez &&&&&{Oromo} & gaz &\checkmark && \checkmark\\
   \multicolumn{1}{c}{}  & \multirow{1}{*}{Hausa} & hau &\checkmark&&&&{Somali} & som &\checkmark&&&& \\
    \multicolumn{1}{c}{}  & \multirow{1}{*} {Tachelhit} & shi & &&&& {Tamazight} & tzm &&&&&  \\ 
 \multicolumn{1}{c}{}  & \multirow{1}{*}{Tamajaq} & ttq &&&&& {Tamajaq} & ttq &&&&& \\ 
 \multicolumn{1}{c}{}  & \multirow{1}{*}{Wolaytta} & wal &&&&& {Tumbuka} & tum &&&&& \\ 
  \multicolumn{1}{c}{\multirow{-6}{*}{\rotatebox[origin=c]{90}{\textbf{\small Afro-Asiatic}}}}  & \multirow{1}{*}{Arabic} & ara &&&&& {Arabic Sudanese Spoken} & apd &&&&& \\ 
 \multicolumn{1}{c}{} & Tigr\'{e}   & tig  &  &  &&& {Tigrinya} & tir & && \\\hline

 \multicolumn{1}{c}{\multirow{-1}{*}{\rotatebox[origin=c]{90}{\textbf{\small A.}}}} & {Malagasy} & plt & & && &{}&{} &&&\\\hline

\multicolumn{1}{c}{}  & \multirow{1}{*}{Akoose} & bss & \checkmark&\checkmark&\checkmark && {Akan} & aka &\checkmark&&\checkmark&& \checkmark \\
\multicolumn{1}{c}{}  & \multirow{1}{*}{Akoose} & bss & \checkmark&\checkmark&\checkmark && {Bambara} & bam  &\checkmark&&\checkmark&&  \\
\multicolumn{1}{c}{}  & \multirow{1}{*}{Bassa} & bsq & \checkmark&\checkmark&\checkmark && {Bemba} & bem &\checkmark&\checkmark&&&  \\

\multicolumn{1}{c}{}  & \multirow{1}{*}{Chitonga } & toi & \checkmark&&&& {Chichewa} & nya & \checkmark&& \\
\multicolumn{1}{c}{}  & \multirow{1}{*}{Dagaare} & dga & \checkmark&&\checkmark&\checkmark& {Dagbani} & dag & \checkmark&& \\ 
\multicolumn{1}{c}{}  & \multirow{1}{*}{Ewondo} & ewo & \checkmark&\checkmark&&& {Farefare} & gur & \checkmark&&\checkmark& \\ 

\multicolumn{1}{c}{}  & \multirow{1}{*}{Fang } & fan & \checkmark&\checkmark&&& {Efik} & efi & \checkmark&\checkmark& \\ 
\multicolumn{1}{c}{}  & \multirow{1}{*}{\'{E}w\'{e}} & ewe & \checkmark&\checkmark&&\checkmark& {Edo} & bin & \checkmark&\checkmark& \\ 
\multicolumn{1}{c}{}  & \multirow{1}{*}{Esan} & ish & \checkmark&&&& {Dangme} & ada & \checkmark&& \\ 
\multicolumn{1}{c}{}  & \multirow{1}{*}{Fulah} & ful &&&&& {Fulfulde} & fuv & && \\
\multicolumn{1}{c}{}  & \multirow{1}{*}{Limba} & lma &&&&& {Fulfulde} & fuv & && \\
\multicolumn{1}{c}{}  & \multirow{1}{*}{Ga} & gaa &\checkmark&&\checkmark&& {Igala} & igl &\checkmark &\checkmark&\checkmark \\
\multicolumn{1}{c}{}  & \multirow{1}{*}{Kabiy\`{e}} & kbp & \checkmark& \checkmark& \checkmark&& {Kpelle } & xpe &\checkmark &\checkmark& \checkmark\\ 
\multicolumn{1}{c}{}  & \multirow{1}{*}{Kikuyu} & kik & \checkmark&& \checkmark&& {Kinyarwanda} & kin &\checkmark && \checkmark\\ 
\multicolumn{1}{c}{}  & \multirow{1}{*}{Igbo} & ibo &\checkmark&\checkmark&\checkmark&& {Mbukushu} & mhw & && \\
\multicolumn{1}{c}{}  & \multirow{1}{*}{Mampruli} & maw &&&&& {Ndonga} & ndo& && \\
\multicolumn{1}{c}{}  & \multirow{1}{*}{Medumba} & byv &\checkmark&\checkmark&&& {Mende} & men &\checkmark && \\
\multicolumn{1}{c}{}  & \multirow{1}{*}{Lunda} & lun &&&&& {Ndebele} & nbl & && \\
\multicolumn{1}{c}{}  & \multirow{1}{*}{Jula} & dyu &&&&& {Kamba} & kam &&& \\
\multicolumn{1}{c}{}  & \multirow{1}{*}{Kabiy\`{e}} & kbp &&&&& {Isoko} & iso &&& \\
\multicolumn{1}{c}{}  & \multirow{1}{*}{Kaonde} & kqn &&&&& {Karaboro} & kza &&& \\
\multicolumn{1}{c}{}  & \multirow{1}{*}{Kimbudu} & kmb &&&&& {Fante} & aka &\checkmark&&\checkmark \\
\multicolumn{1}{c}{}  & \multirow{1}{*}{Kwanyama} & kua &&&&& {Luganda} & lug &\checkmark&& \\
\multicolumn{1}{c}{}  & \multirow{1}{*}{Kongo} & kwy &\checkmark&\checkmark&\checkmark&& {Kwangali} & kwn &&& \\
\multicolumn{1}{c}{\multirow{-10}{*}{\rotatebox[origin=c]{90}{\textbf{\small Niger Congo}}}}  & \multirow{1}{*}{Twi} & aka &\checkmark&\checkmark&\checkmark&& {Chitumbuka} & kwn &\checkmark&& \\

\multicolumn{1}{c}{}  & \multirow{1}{*}{Tswa} & tsc &&&&& {Tshiluba} &lua &&& \\
\multicolumn{1}{c}{}  & \multirow{1}{*}{Tsonga} & tso &&&&& {Zama} &xuu &&& \\
\multicolumn{1}{c}{}  & \multirow{1}{*}{Limba} & lma &\checkmark&\checkmark&\checkmark&& {Lukpa} & dop &&& \\
\multicolumn{1}{c}{}  & \multirow{1}{*}{Pular} & fuf &&&&& {Kissi} & kqs &\checkmark && \\
\multicolumn{1}{c}{}  & \multirow{1}{*}{Rundi} & run &&&&& {Setswana} & tsn &\checkmark && \\
\multicolumn{1}{c}{}  & \multirow{1}{*}{Shona} & sna &\checkmark&&\checkmark&& {Swahili Congo} & swc & && \\ 
\multicolumn{1}{c}{}  & \multirow{1}{*}{Swati } & ssw & \checkmark&&&& {Swahili} & swh & && \\
\multicolumn{1}{c}{}  & \multirow{1}{*}{Swahili} & swa & &&&& {Sepedi} & nso & && \\
\multicolumn{1}{c}{}  & \multirow{1}{*}{Sesotho} & sot &\checkmark&&\checkmark&& {Tsonga} & tso & && \\
\multicolumn{1}{c}{}  & \multirow{1}{*}{Themne} & tem &&&&& {Comorian Ngazidja} & zdj & && \\
\multicolumn{1}{c}{}  & \multirow{1}{*}{Urhobo} & urh &\checkmark&&\checkmark&\checkmark& {Tshiluba} & lua & && \\
\multicolumn{1}{c}{}  & \multirow{1}{*}{Venda} & ven &&&&& {Wolof} & wol & &&\checkmark \\
\multicolumn{1}{c}{}  & \multirow{1}{*}{Xhosa} & xho &\checkmark&&&& {Lingala} & lin &\checkmark&\checkmark&\checkmark&&  \\
\multicolumn{1}{c}{}  & \multirow{1}{*}{Yemba} & ybb &&&&& {Yor\`{u}b\'{a}} & yor &\checkmark&\checkmark&\checkmark&\checkmark&  \\
\multicolumn{1}{c}{}  & \multirow{1}{*}{Zande} & zne &&&&& {Mboshi} & mdw &&&&&  \\
 \multicolumn{1}{c}{} & Zulu  & zul &\checkmark&&&& {Tiv} & tiv&\checkmark&&&\\\hline

\multicolumn{1}{c}{}  & \multirow{1}{*}{Kanuri} & knc &\checkmark&\checkmark&\checkmark&& {Dinka} & dik &\checkmark&\checkmark&&&  \\
\multicolumn{1}{c}{\multirow{-2}{*}{\rotatebox[origin=c]{90}{\textbf{\small N.S.}}}}  & \multirow{1}{*}{Kunama} & kun &&&&&{Bari} &bfa  &&&&&  \\
 \multicolumn{1}{c}{} & Luo & luo  & \checkmark & &&& {Dendi}& ddn & \\\hline
 
\multicolumn{1}{c}{}  & \multirow{1}{*}{Afrikaans} & afr &&&&& {English} & eng &&&&&  \\
\multicolumn{1}{c}{}  & \multirow{1}{*}{French} & fra &&&&& {Portuguese} & por &&&&&  \\
\multicolumn{1}{c}{\multirow{-3}{*}{\rotatebox[origin=c]{90}{\textbf{\small I.E.}}}} & {Spanish} & spa  &  & &&& Urdu &urd\\\hline

\multicolumn{1}{c}{}  & \multirow{1}{*}{Kituba} & ktu &&&&& Juba Arabic& pga &&&&&  \\
\multicolumn{1}{c}{\multirow{-2}{*}{\rotatebox[origin=c]{90}{\textbf{\small Creole}}}}  & \multirow{1}{*}{Seychelles French Creole} & crs &&&&& Sango& sag &\checkmark&\checkmark&&&  \\
\multicolumn{1}{c}{} & {Nigerian Pidgin} & pcm  & & && &{Kabyverdianu}&{kea} &&&\\\bottomrule

\end{tabular}}
\end{adjustbox}
\caption{List of Languages, language codes and typology of the languages presented in this paper across 6 language families: Austronesian (A.), Nilo-Saharan (N.S.), and Indo-European (I.E.). The checkmarks are added to each language to indicate the presence of the corresponding feature.}\label{tab:lang-info}
\end{table*}

\section{The Language Situation in Africa}
In Table \ref{tab:lang-use} we list the status for different languages in Africa. This table was created using information available on ethnologue \cite{ethnologue} for each African country. The national, regional, educational and Indigenous languages are presented as it applies to each country. We present all African countries including those not officially recognized in this list. To the best of our knowledge, this list is a true representation of the status of languages used in Africa. 

All African countries, except Ethiopia and Liberia were colonized, with most gaining independence between the 1950s and the 1970s. The colonialist came from different parts of Europe and adopted different language policies which seem to play an important role in the language policies adopted by different African countries today. Although economics, politics, and globalization also play a crucial role. All colonialists interacted derogatorily with Indigenous languages and often referred to them as \textit{vernaculars}. Although, the British colonialists allowed Indigenous languages in their territories if desired. The French, Spanish, and Portuguese on the other hand did not tolerate any Indigenous languages in public. Despite the British's tolerance for Indigenous languages, Indigenous languages were allowed only in early childhood education and Indigenous languages where prohibited after the 4th year in elementary school~\cite{Williams+2013+68+87, ouane2010and}. 

From Table~\ref{tab:lang-use}, it is evident that colonial languages have retained their official status in many African countries till date~\cite{khejeri2014teachers}. Foreign languages are dominantly used in education, and most official government functions, even in countries where Indigenous languages have official status~\cite{banda2009critical}. According to~\citet{ouane2010and}, only $25\%$ of the languages used in secondary education and $5\%$ of the languages in higher education are African. This is despite the known benefits of using Indigenous languages in Education and minority language development~\cite{buhmann2008mother, TRUDELL2005237, Williams+2013+68+87, bull1955use}. In cases where policy favours the official use of Indigenous languages, some governments have shown a lack of political will to implement these policies~\cite{williams2011language}. The current linguistic situation thus seem to be one of convenience rather than one from well developed language policies.

Despite a few dissenting voices who argue that the use of several mother tongues will accentuate inter-tribal conflict \cite{khejeri2014teachers}, the general consensus is that preserving language diversity through policies that encourage multilingualism are most desirable. Developing a truly multilingual language policy for Africa will certainly be challenging \cite{ouane2010and}, but will be most beneficial even to the progress of NLP on the African continent.

\begin{table*}[h!]
\small
\centering
\begin{adjustbox}{max width=\textwidth}
\renewcommand{\arraystretch}{1.1}
{
        \begin{tabular}{>{}clllllll}
        \toprule

      \textbf { \small Region}    &\textbf { \small Country }   &\textbf {\small Lang(s)} & \textbf{ \small Ind.} & \textbf{ \small National} & \textbf{ \small Regional} & \textbf{ \small Educational}\\   \midrule
      
 \multicolumn{1}{c}{}  & \multirow{1}{*}{\textbf{Burundi}} & 4 & 2 & run & & & \\ 
 \multicolumn{1}{c}{}  & \multirow{1}{*}{\textbf{Comoros}} & 7 & 2 &  fra, ara & zdj & & \\ 
 \multicolumn{1}{c}{}  & \multirow{1}{*}{\textbf{Djibouti}} & 5 & 2 & fra, ara & & & \\ 
 \multicolumn{1}{c}{}  &   \multirow{1}{*}{\textbf{Eritrea}} & 15 & 9 & ara & & kun, tig\\ 
 \multicolumn{1}{c}{}  &   \multirow{1}{*}{\textbf{Ethiopia}} & 91 & 87 & amh &aar, gaz, som, tir & 31 Ind. and 1 foreign& \\ 
 \multicolumn{1}{c}{}  &   \multirow{1}{*}{\textbf{Kenya}} & 68 & 61 & eng, swa & & 11 Ind. & \\ 
 \multicolumn{1}{c}{}  &   \multirow{1}{*}{\textbf{Madagascar}} & 14 & 12 & fra, mlg (higher ed.) & &\\ 
 \multicolumn{1}{c}{}  &   \multirow{1}{*}{\textbf{Malawi}} & 17 & 13 & eng & & tum \\ 
 
 \multicolumn{1}{c}{}  &   \multirow{1}{*}{\textbf{Mauritius}} & 9 & 2 & eng, fra & & urd\\ 
 
  \multicolumn{1}{c}{}  &   \multirow{1}{*}{\textbf{Mayotte}} & 4 & 2 & fra & &\\ 
 
  \multicolumn{1}{c}{}  &   \multirow{1}{*}{\textbf{Mozambique}} & 44 & 42 & por & & \\ 
 
  \multicolumn{1}{c}{}  &   \multirow{1}{*}{\textbf{Reunion}} & 3 & 1 & fra & &\\ 
  \multicolumn{1}{c}{}  &   \multirow{1}{*}{\textbf{Rwanda}} & 4 & 2 & eng, fra, kin & & \\ 
 
  \multicolumn{1}{c}{}  &   \multirow{1}{*}{\textbf{Seychelles}} & 3 & 1 & eng, fra, crs & & \\ 
 
  \multicolumn{1}{c}{}  &   \multirow{1}{*}{\textbf{Somalia}} & 13 & 10 & ara, som & & eng \\ 
 
   \multicolumn{1}{c}{\multirow{-13}{*}{\rotatebox[origin=c]{90}{\textbf{\small East Africa}}}}  &   \multirow{1}{*}{\textbf{South Sudan}} & 70 & 59 & eng & pga, zne, apd, bfa & 8 Ind.  \\ 
 
  \multicolumn{1}{c}{}  &   \multirow{1}{*}{\textbf{Tanzania}} & 126 & 118 & swa & & eng and swa\\ 
 
  \multicolumn{1}{c}{}  &   \multirow{1}{*}{\textbf{Uganda}} & 44 & 41 & eng, swa & & 2 Ind. , 1 non-Ind..  \\ 
 
  \multicolumn{1}{c}{}  &   \multirow{1}{*}{\textbf{Zambia}} & 46 & 37 & eng & 3 Ind. & 4 Ind., 1 non-Ind.\\ 
 
 \multicolumn{1}{c}{}  & \textbf{Zimbabwe} & 22 & 16 & eng & & 2 Ind, 2 South African\\ 

    \hline
 \multicolumn{1}{c}{}  & \multirow{1}{*}{\textbf{Angola}} & 46 & 41 & por & &\\ 
 \multicolumn{1}{c}{}  & \multirow{1}{*}{\textbf{Cameroon}} & 275 & 271 & eng, fra & & \\ 
 \multicolumn{1}{c}{}  & \multirow{1}{*}{\textbf{Central Afr. Rep.}} & 75 & 65 & eng, sag& & \\ 
 \multicolumn{1}{c}{}  & \multirow{1}{*}{\textbf{Chad}} & 129 & 123 & fra, ara & & \\ 
 \multicolumn{1}{c}{}  & \multirow{1}{*}{\textbf{Congo}} & 66 & 55 & fra & & \\ 
 \multicolumn{1}{c}{}  & \multirow{1}{*}{\textbf{Dem. Rep. of Congo}} & 214 & 209 & fra & & 2 Ind.\\ 
 \multicolumn{1}{c}{}  & \multirow{1}{*}{\textbf{Equatorial Guinea}} & 15 & 12 & spa& & 2 foreign  \\ 
 \multicolumn{1}{c}{}  & \multirow{1}{*}{\textbf{Gabon}} & 43 & 40 & fra & & \\ 
 \multicolumn{1}{c}{\multirow{-9}{*}{\rotatebox[origin=c]{90}{\textbf{ \small Middle Africa}}}}  &\textbf{Sao Tome e Principe}  & 7 & 3 & por & & 1 foreign \\ 
    \hline
 \multicolumn{1}{c}{}  & \multirow{1}{*}{\textbf{Algeria}} & 19 & 14 & ara, kab& & 1 foreign \\ 
 \multicolumn{1}{c}{}  & \multirow{1}{*}{\textbf{Egypt}} & 19 & 9 & ara & &  \\ 
 \multicolumn{1}{c}{}  & \multirow{1}{*}{\textbf{Libya}   } & 9 & 8 & ara & &  \\ 
 \multicolumn{1}{c}{}  & \multirow{1}{*}{\textbf{Morocco}   } & 15 & 10 & ara, tzm & &  \\ 
 \multicolumn{1}{c}{}  & \multirow{1}{*}{\textbf{Sudan}   } & 75 & 70 & ara, eng & & \\ 
 \multicolumn{1}{c}{\multirow{-6}{*}{\rotatebox[origin=c]{90}{\textbf{North Africa}}}}  & \textbf{Western Sahara} & 4 & 2 & ara & & \\ 
    \hline
 \multicolumn{1}{c}{}  & \multirow{1}{*}{\textbf{Botswana}} & 31 & 26 & eng and tsn& & 1 foreign \\ 
 \multicolumn{1}{c}{}  & \multirow{1}{*}{\textbf{Eswatini}} & 5 & 1 & eng, ssw  & & \\ 
 \multicolumn{1}{c}{}  & \multirow{1}{*}{\textbf{Lesotho}} & 5 & 3 & eng and sot & & \\ 
 \multicolumn{1}{c}{}  & \multirow{1}{*}{\textbf{Namibia}} & 28 & 23 & eng & & 3 foreign, 6 Ind.\\ 

{\multirow{-4}{*}{\rotatebox[origin=c]{90}{\textbf{\small South Africa}}}}  & \textbf{\small\textbf{South Africa}}& & & eng, afr, nbl, tsn, \\
{\multirow{-6}{*}{\rotatebox[origin=c]{90}{\textbf{\small }}}}  & & & &  nso, sot, ssw,  \\
{\multirow{-6}{*}{\rotatebox[origin=c]{90}{\textbf{}}}} &  & 31 & 20 &  tso, ven, xho, zul & & 1 foreign 
\\ 

   \hline
 \multicolumn{1}{c}{}  & \multirow{1}{*}{\textbf{Benin}} & 55 & 50 & fra & &\\ 
 \multicolumn{1}{c}{}  & \multirow{1}{*}{\textbf{Burkina Faso}} & 71 & 66 & fra && \\ 
 \multicolumn{1}{c}{}  & \multirow{1}{*}{\textbf{Cape Verde Islands}} & 2 & 1 & kea, por & &  \\ 
 \multicolumn{1}{c}{}  & \multirow{1}{*}{\textbf{Cote d'ivoire}} & 87 & 75 & fra & & 1 foreign\\ 
 \multicolumn{1}{c}{}  & \multirow{1}{*}{\textbf{Gambia}} & 11 & 7 & eng & & \\ 
 \multicolumn{1}{c}{}  & \multirow{1}{*}{\textbf{Ghana}} & 83 & 73 & eng & gur, maw & 5 Ind.\\ 
 \multicolumn{1}{c}{}  & \multirow{1}{*}{\textbf{Guinea}} & 37 & 35 & fra & fuf& 3 Ind.\\ 
 \multicolumn{1}{c}{}  & \multirow{1}{*}{\textbf{Guinea-Bissau}} & 23 & 18 & por & &\\ 
\multicolumn{1}{c}{}  & \multirow{1}{*}{\textbf{Liberia}} & 31 & 27 & eng & & \\ 
 \multicolumn{1}{c}{}  & \multirow{1}{*}{\textbf{Mali}} & 68 & 63 & fra  & & 5 Ind., 1 foreign\\ 
 \multicolumn{1}{c}{}  & \multirow{1}{*}{\textbf{Mauritania}} & 7 & 5 & ara & & \\ 
 \multicolumn{1}{c}{}  & \multirow{1}{*}{\textbf{Niger}} & 23 & 19 & fra & & 2 Ind., 1 foreign \\ 
 \multicolumn{1}{c}{}  & \multirow{1}{*}{\textbf{Nigeria}} & 522 & 512 & eng & hau, ibo, yor & 5 Ind.\\ 
 \multicolumn{1}{c}{}  & \multirow{1}{*}{\textbf{Saint Helena}} & 1 & 0 & eng && \\ 
 \multicolumn{1}{c}{\multirow{-13}{*}{\rotatebox[origin=c]{90}{\textbf{\small West Africa}}}}  & \multirow{1}{*}{\textbf{Senegal}} & 39 & 31 & fra & & \\ 
 \multicolumn{1}{c}{}  & \multirow{1}{*}{\textbf{Sierra Leone}} & 24 & 19 & eng & lma, men, tem& \\ 
 \multicolumn{1}{c}{}  & \textbf{Togo} & 44 & 40 & fra & & \\ 
   \bottomrule
\end{tabular}}
\end{adjustbox}
\caption{A statistics of the language use in Africa computed from \cite{ethnologue}. For each country, we show the number of languages (lang) reported, the number of Indigenous languages spoken in the country (Ind.), the national languages, the regional languages, and the educational languages. }\label{tab:lang-use}
\end{table*}

\section{Communities}\label{subsec:community}
Many communities contribute significantly to the development of NLP for African languages. We list some of them below. 
\href{https://www.masakhane.io/}{Masakhane} aims to build an active community geared at creating resources that are truly representative of African culture, facilitating collaborations to develop African NLP and lowering the barriers for NLP participation. They achieve this by having an active slack channel that fosters interaction between stakeholders, organizing workshops, creating easy to use google colab notebooks among several other initiatives. Masakhane so far has over 1,000 members. 

\href{https://blackinai.github.io/#/}{Black in AI} is an organization that focuses on increasing the presence, inclusion, and visibility of black people in artificial intelligence. They achieve this objective through advocacy, mentorship, and facilitating collaborations. Although BlackinAI encompasses black people beyond the African continent, and they do not specifically restrict their operations to African languages, it is a great community for collaborations.

\href{https://deeplearningindaba.com/about/our-mission/}{Deep Learning Indaba} is an organisation whose mission is to strengthen machine learning and artificial intelligence in Africa by enabling Africans to be active shapers and owners of AI technologies. Deep Learning Indaba which was inaugurated in 2017 organizes an annual Deep Learning Indaba retreat for teachings and practical sessions on AI. They also provide mentorship programs and grants (the IndabaX) that fund AI gatherings in 26 African countries with plans underway to include more countries. This is in addition to awards for the application of AI to an African problem, for excellence in research in tertiary African institutions, and for services to the machine learning community in Africa- Kambule, Maathai, and Umuntu awards respectively. These programmes aim to build a sustainable pan-African community of AI expertise, create local leadership in AI in every country across the continent, and recognise excellence in research and application of AI technologies, respectively.

\href{https://www.k4all.org/}{Knowledge 4 All} Foundation Ltd (K4A) pioneers machine learning methods of pattern analysis, statistical modeling, and computational learning and transforms these into technologies for large scale applications in open education. They organize symposiums, summer schools, workshops, colloquiums, and conferences. They also provide fellowship to develop datasets and strengthen capacities and innovation potential for low resource African languages under the international development program. They have developed resources for Ewe, Fongbe, Y{o}r\`{u}b\'{a}, Chichewa, Wolof, Kiswahili, Tunisian Arabizi, Twi, and Luganda. They also various competitions to develop or improve methods for NLP of African languages.

\href{https://zindi.africa/}{Zindi} hosts a large community of African data scientists and facilitates collaborations between data scientists and organizations. They provide a place to learn, improve skills and find a job. They also organize competitions for data collection tasks and developing NLP models for various African languages. 

\href{http://www.alt-i.org/}{ALTI} is one of the pioneering NLP communities in Africa. They focus on making computers usable in African languages and develop and grow human talent that take African Languages into the information age. They also provide a hub were NLP enthusiasts can be mentored for NLP work in African languages.

Different organization provide funding for NLP research. Some of these organizations are presented in Table \ref{tab:funding}.

\begin{table}[h!]
\small 
\centering
\begin{tabular}{>{}c|l r}
 \toprule
\multicolumn{1}{c}{}  &\textbf{Organization} & \textbf{Type} \\ 
 \midrule
\multicolumn{1}{c}{}  &\href{https://www.google.org/}{Google} & Industry  \\
\multicolumn{1}{c}{}  &\href{https://www.microsoft.com/en-ca/about/}{Microsoft} & Industry  \\
\multicolumn{1}{c}{}  &\href{https://www.rockefellerfoundation.org/commitment/innovation/}{The Rockefeller Foundation} & Foundation   \\
\multicolumn{1}{c}{}  &\href{https://www.rockefellerfoundation.org/commitment/innovation/}{FAIR Forward} & Government  \\

\multicolumn{1}{c}{}  & \href{https://lacunafund.org/}{Lacuna Funds} & NGO\\
\multicolumn{1}{c}{}  & \href{https://www.k4all.org/}{Knowledge 4 All} & Research \\
\multicolumn{1}{c}{}  & \href{https://www.idrc.ca/en}{IDRC} & Research  \\

 \bottomrule
\end{tabular}
\caption{Some funding Organization for African NLP including Non-Governmental organizations (NGO)s}\label{tab:funding}
\end{table}

\section{Workshops}\label{subsec:workshops}
The \textit{\textbf{AfricanNLP}} workshop has run annually alongside ICLR and EACL in $2020$ and $2021$ respectively. In $2020$, $32$ papers were presented while in $2021$, $40$ papers describing different systems were accepted. Currently, papers submitted are non-archival, giving authors the opportunity to submit the papers to other venues. \textit{\textbf{BlackInAI}} has organized yearly workshops co-located with Neural Information Processing Systems (NeurIPS) since $2017$. Audience is composed of researchers who self-identify as Black and often has many works related to African languages.

\section{Resources}\label{subsec:resources}
\subsection{Labelled Resources}
Majority of labelled corpora is developed as part of the development process of many NLP tasks. This is due to a lack of readily available labelled corpora for many NLP tasks. Labelled corpora has been developed for MT \cite{adelani2021menyo, nekoto2020participatory, tapo2020neural, dossou2020ffr, ezeani2020igbo, hadgu2020evaluating}, classification \cite{niyongabo2020kinnews, fourati2020tunizi, oyewusi2020semantic}, automatic spelling correction \cite{gezmu2018portable}, morphological segmentation \cite{outahajala2016using, mott2020morphological}, NER \cite{ifeoluwa2021masakhaner, hedderich2020survey}, diacritic restoration \cite{orife2020improving, asahiah2017restoring}, automatic speech recognition (ASR) \cite{dossou2021okwugb, tachbelie2020analysis}, and speech translation \cite{DBLP:journals/corr/abs-1710-03501}. A summary of labelled corpora can be found in Table \ref{tab:resources}. 

A few of the labelled corpora are developed by trained linguists and language experts \cite{strassel2016lorelei, adebaratranslating} while others are collected by native speakers \cite{ifeoluwa2021masakhaner, adelani2021menyo, nekoto2020participatory}. Furthermore, evaluation is often done using automatic metrics that measure model performance rather than data quality or inter-annotator agreement \cite{outahajala2016using}. Data is also often labelled on the assumption that the data has been proofread \cite{gezmu2018portable}, while the procedure for developing the dataset is often not discussed.  It is important to mention here that we advocate that trained linguists or language experts, particularly those trained in African languages, be involved in data collection or curation activities for African languages. This is because of the linguistic situation in Africa and the literacy levels in African languages which have been discussed in this paper.

\subsection{Unlabelled Corpora}
Unlabelled corpora seem to be the bulk of available data for African languages. Most corpora are crawled from the web as part of multilingual corpora development efforts like JW300 \cite{agic-vulic-2019-jw300}, ParaCrawl \cite{banon-etal-2020-paracrawl, espla2019paracrawl}, WikiMatrix \cite{schwenk2019wikimatrix}, OSCAR \cite{suarez2020monolingual}, mC4 \cite{xue-etal-2021-mt5}, CCAligned \cite{elkishky_ccaligned_2020}, wikiAnn \cite{pan2017cross}. We provide a summary of unlabelled corpora in Table \ref{tab:resources}.

\subsection{Crosslingual Tools} 
Pre-trained models like BERT \cite{devlin2018bert}, ELMo \cite{peters2018deep}, Roberta \cite{liu2019roberta}, GPT \cite{radford2018improving, radford2019language, brown2020language}, BART \cite{lewis2019bart} have advanced the state of the art in a wide variety of tasks, suggesting that these models acquire valuable, generalizable linguistic information during the pre-training process. However, training language-specific models is possible for only a few languages which have large amounts of data. A popular alternative has been multilingual language models (MLM) such as mBERT \cite{devlin2018bert}, XML-R \cite{conneau2019unsupervised}, MT5 \cite{xue-etal-2021-mt5}, mBART \cite{liu2020multilingual} and many others. MLMs are trained on large amounts of unlabelled data from multiple languages so that low resource languages may benefit from shared vocabulary and other linguistic information from high resource languages and other similar languages in the MLM. Very few MLMs have representations for African languages and many of those available are trained with noisy data \cite{adelani2021effect, alabi2020massive, caswell2021quality} which may affect downstream tasks. We provide information about crosslingual tools in Table \ref{tab:cross-lingual} and other NLP models in Table {tab:modelresources}.

\subsection{Raw Data}
Blog sites, online newspapers, Wikipedia, Jehovah's Witness website are some sources of raw data for African languages. We provide details in Table \ref{tab:raw-data} and Table \ref{tab:resources}.


\begin{table}[h!]
\small
\centering
\begin{tabular}{@{}lll@{}}
\toprule
\multicolumn{1}{c}{\textbf{Country}} & \multicolumn{1}{c}{\textbf{Site}}                                                             & \multicolumn{1}{c}{\textbf{Language}} \\ \midrule
Ethiopia                             & \href{https://www.addisadmassnews.com/}{Addisadmassnews}                   & amh                               \\
Ethiopia                             & \href{http://ethiopianreporter.com/}{Ethiopian Reporter}                   & amh and eng                   \\
Lesotho                              & \href{https://mosotho.co.ls/}{Mosotho}                                     & sot                               \\
Namibia                              & \href{https://www.republikein.com.na/}{Republikein}                        & afr                              \\
Nigeria                              & \href{https://hausa.premiumtimesng.com/}{Premiumtimes}                     & hau                                 \\
Nigeria                              & \href{https://hausa.leadership.ng/}{Leadership}                            & hau                                 \\
Nigeria                              & \href{https://hausa.legit.ng/}{Hausa Legit}                                & hau                                 \\
Nigeria                              & \href{https://aminiya.dailytrust.com/}{Aminiya}                            & hau                                 \\
Nigeria                              & \href{https://www.igboradio.com/}{Igbo Radio}                              & ibo                                  \\
Nigeria                              & \href{https://kaoditaa.com/}{Kaoditaa}                                     & ibo                                  \\
Nigeria                              & \href{http://iroyinowuro.com.ng/}{Iroyin Owuro}                            & yor                                \\
Somalia                              & \href{https://boramanews.com/}{Boramanews}                                 & som                               \\
Somalia                              & \href{https://www.caasimada.net/}{Caasimada}                               & som                               \\
Somalia                              & \href{https://horseedmedia.net/}{Horseedmedia}                             & som                               \\
Somalia                              & \href{http://www.idalenews.com/}{Idalenews}                                & eng and som                   \\
Somalia                              & \href{http://markacadeey.com/}{Markacadeey}                                & eng and som                   \\
Somalia                              & \href{https://ogaden.com/}{Ogaden}                                         & eng and som                   \\
Somalia                              & \href{https://puntlandpost.net/}{Puntlandpost}                             & eng and som                   \\
Somalia                              & \href{https://qarannews.com/}{PQarannews}                                  & eng and som                   \\
Somalia                              & \href{https://shabellemedia.com/}{Shabellemedia}                           & eng and som                   \\
Somalia                              & \href{http://www.simbanews.net/}{Simbanews}                                & eng and som                   \\
Somalia                              & \href{https://www.togaherer.com/}{Togaherer}                               & eng and som                   \\
Somalia                              & \href{https://www.waagacusub.com/}{Waagacusub}                             & eng and som                   \\
Somaliland                           & \href{https://www.dhamaysnews.com/}{Dhamays news}                          & som                               \\
Somaliland                           & \href{https://goobjoog.com/category/wararka-dalka-2/somaliland/}{Goobjoog} & som                               \\
Somaliland                           & \href{http://haatuf.net/}{Haatuf}                                          & som                               \\
Somaliland                           & \href{https://maandeeq.net/}{Maandeq}                                      & som                               \\
Somaliland                           & \href{https://qorilugudnews24.com/}{Qorilugudnews}                         & som                               \\
Somaliland                           & \href{https://somalilandpost.net/}{Somalilandpost}                         & eng and som                   \\
South Africa                         & \href{https://www.netwerk24.com/netwerk24/za/beeld}{Netwerk24}             & afr                            \\
South Africa                         & \href{https://www.netwerk24.com/huisgenoot}{Huisgenoot}                    & afr                            \\
South Africa                         & \href{ https://www.dievryburger.co.za/}{Dievryburger}                      & afr                            \\
South Africa                         & \href{https://www.isolezwe.co.za/}{Isolezwe}                               & zul                                  \\
Tanzania                             & \href{https://www.mwananchi.co.tz/}{Mwananchi}                             & swh                               \\
Tanzania                             & \href{https://www.ippmedia.com/sw/nipashe}{Nipashe}                        & swh                               \\
Tanzania                             & \href{https://www.ippmedia.com/sw/nipashe-jumapili}{Nipashe-Jumpaili}      & swh                               \\
Uganda                               & \href{http://www.bukedde.co.ug/}{Bukedde}                                  & lug                               \\
Zimbabwe                             & \href{https://www.kwayedza.co.zw/}{Kwayedza}                               & sna                                 \\
Zimbabwe                             & \href{https://www.umthunywa.co.zw/}{Umthunywa}                             & nbl                               \\ \bottomrule
\end{tabular}
\caption{Newspapers in Indigenous languages of Africa.} \label{tab:raw-data}
\end{table}

\begin{figure*}[h!]
\centering
\begin{adjustbox}{width=14cm}
\renewcommand{\arraystretch}{1.5}
{
  \includegraphics[width=10cm,height=8cm]{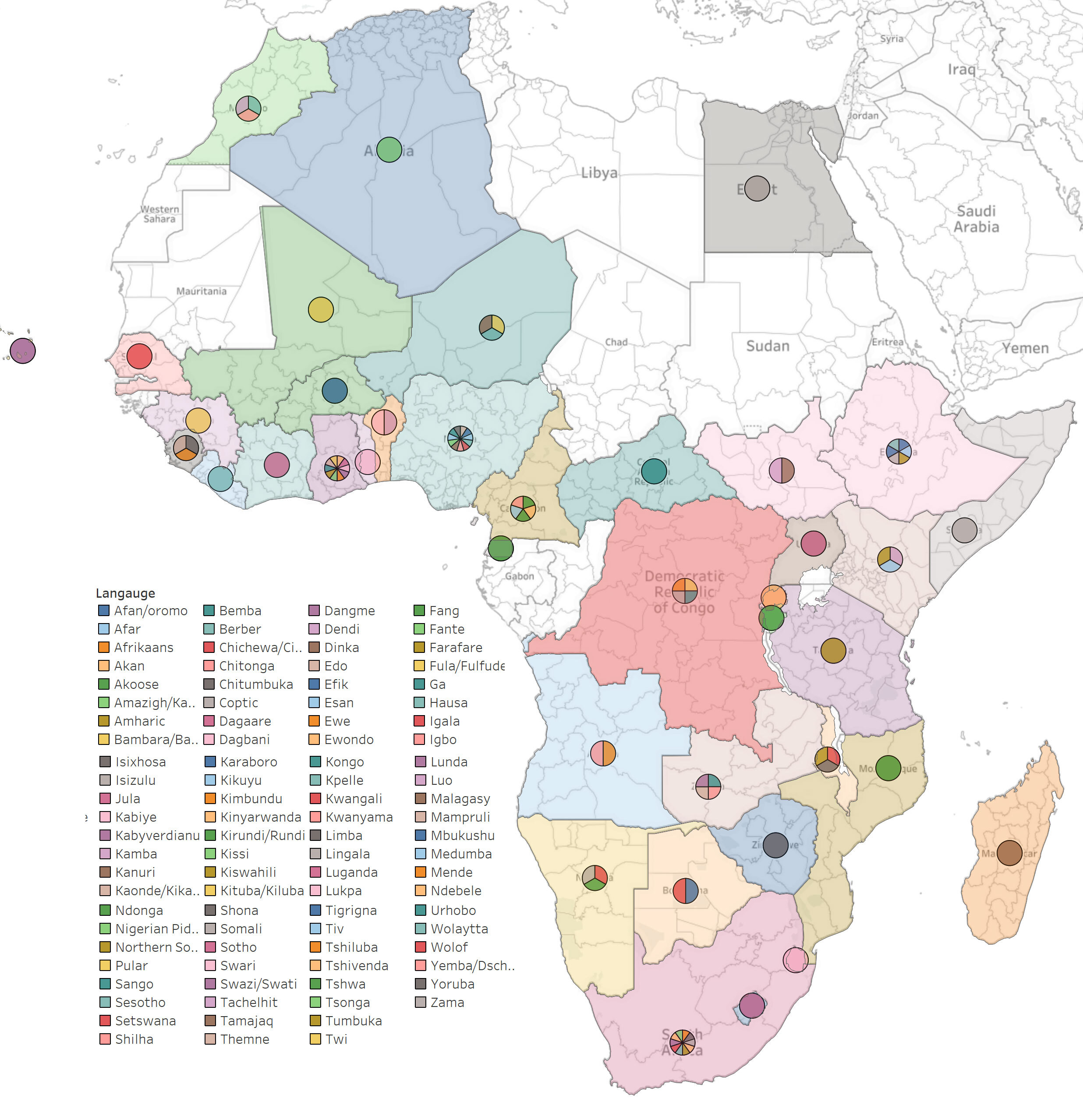}}
  \end{adjustbox}
  \caption{A high quality (bigger) version of the African languages map provided in this paper.}
\label{fig:quality_our_map}
\end{figure*}

\begin{table*}[h!]
\centering
\small
\begin{adjustbox}{max width=\textwidth}
\renewcommand{\arraystretch}{1.15}
\begin{tabular}{llll}
\toprule
\textbf{\begin{tabular}[c]{@{}l@{}}Num Score\end{tabular}} &              & \textbf{\begin{tabular}[c]{@{}l@{}}Most Extensive Grammar Description Type\end{tabular}}               & \textbf{\# Languages} \\ \hline
$5$                                                             & long grammar & \begin{tabular}[c]{@{}l@{}}extensive description of most features of the grammar\\ $\approx$$300+$ pages\end{tabular} & \quad $411$  \quad  $18.9\% $        \\

$4$                                                            & grammar & \begin{tabular}[c]{@{}l@{}}a description of most features of the grammar\\ $\approx$$150$ pages\end{tabular} & \quad $243$    \quad $11.1\%$         \\

$3$                                                            & grammar sketch & \begin{tabular}[c]{@{}l@{}}a less extensive description of many features of the grammar\\ $\approx$$50$ pages\end{tabular} & \quad $562$  \quad  $25.9\%$        \\

$2$                                                            & specific feature & \begin{tabular}[c]{@{}l@{}}a description of some features of the grammar\\ (i.e noun class system, verb morphology, etc)\end{tabular} & \quad $157$  \quad  $7.2\% $        \\

$2 $                                                           & phonology & \begin{tabular}[c]{@{}l@{}}a description of the sound inventory\\ using minimal pairs\end{tabular} & \quad $82$ \quad   $3.7\% $        \\

$2 $                                                           & dictionary & \begin{tabular}[c]{@{}l@{}}$\approx$$75+$ pages\end{tabular} & \quad $53$  \quad  $2.4\% $        \\

$2 $                                                           & text & \begin{tabular}[c]{@{}l@{}}text material\end{tabular} & \quad$13$  \quad  $0.5\% $        \\

$1 $                                                           & wordlist & \begin{tabular}[c]{@{}l@{}}$\approx$$100-200$ words\end{tabular} & \quad $13$ \quad   $0.5\%$        \\

$0 $                                                           & minimal & \begin{tabular}[c]{@{}l@{}}a small number of morphemes \end{tabular} & \quad $124$  \quad  $5.7\% $        \\

$0 $                                                           & overview & \begin{tabular}[c]{@{}l@{}}document with meta-information about the language \\ (i.e where spoken, non-intelligibility to other languages etc) \end{tabular} & \quad $48$  \quad  $2.2\% $        \\
\hline
&  &\quad\quad\quad\quad\quad\quad\quad\quad\quad\quad\quad\quad\quad\quad\quad\quad\quad\quad\quad\quad\quad\quad \textbf{Total:}&  \quad$2,169 $        \\       
\bottomrule
                               
\end{tabular}
 \end{adjustbox}
 \caption{Available linguistic resources for African languages. Adapted from \cite{hock2011world}.} \label{tab:lang_resources}
 
 \end{table*}
 
\begin{table*}[h!]
\centering
\begin{adjustbox}{max width=\textwidth}
\renewcommand{\arraystretch}{1.15}
\begin{tabular}{p{3cm}p{3cm}p{8cm}} 
\toprule
\thead{\textbf{Model}} & \thead{\textbf{Language(s)}}  &  \thead{\textbf{URL}} \\
\midrule
Word embeddings & Twi-Y{o}r\`{u}b\'{a} & \url{https://github.com/ajesujoba/YorubaTwi-Embedding} \\
Okwugbe (ASR) & Igbo-Fon & \url{https://github.com/bonaventuredossou/fonasr} \\
Automatic Diacritic Restoration & Y{o}r\`{u}b\'{a} & \url{https://github.com/Niger-Volta-LTI/yoruba-adr} \\
FFR v.1.1 model & Fon-French & \url{https://github.com/bonaventuredossou/ffr-v1/blob/master/model_train_test/fon_fr.py} \\
Masakhane MT & 30 languages & \url{https://github.com/masakhane-io/masakhane-mt} \\
AfriBERT & Afrikaans & \url{https://github.com/sello-ralethe/AfriBERT} \\
\bottomrule
 \end{tabular}
 \end{adjustbox}
 \caption{A list of available models.}\label{tab:modelresources}
 \end{table*}

\begin{table*}[h!] 
\centering
\begin{adjustbox}{max width=\textwidth}
\renewcommand{\arraystretch}{1.15}
\begin{tabular}{p{3cm}p{10cm}} 
\toprule
\thead{\textbf{Language Model}} & \thead{\textbf{African Languages Represented}}  \\ \midrule
MT5 & afr, nya, mlg hau, ibo, sna, som, sot / nso, swa, xho, yor, zul  \\ 
MBERT & afr, swa, yor \\ 
XLM-R & afr, amh, hau, gaz, som, swa, xho. \\ \bottomrule
 \end{tabular}
 \end{adjustbox}
 \caption{Language models with African languages represented.} \label{tab:cross-lingual}
 \end{table*}

 \begin{table*}[h!]
\small
\begin{adjustbox}{max width=\textwidth}
\renewcommand{\arraystretch}{1.0}
{
        \begin{tabular}{>{}clllll}
        \toprule

      \textbf { }    &\textbf { \small Name }   &\textbf {\small Language(s)} & \textbf{ \small Task.} & \textbf{ \small References} \\    \midrule
 \multicolumn{1}{c}{}  & \multirow{1}{*}{\href{https://github.com/Andrews2017/KINNEWS-and-KIRNEWS-Corpus}{KINNEWS and KIRNEWS Corpus}} & kin, run & POS, NER, Parsing & \cite{niyongabo2020kinnews} \\ 
 \multicolumn{1}{c}{} &  & amh, hau, ibo, kin, lug, \\
 \multicolumn{1}{c}{} & \href{https://github.com/masakhane-io/masakhane-ner}{Masakhane NER} & luo, pcm, swa, wol, yor & NER & \cite{ifeoluwa2021masakhaner} \\ 
 \multicolumn{1}{c}{} & \href{https://git.io/JvHrp and https://git.io/Jv9og}{Nigerian Pidgin Tweets} & pcm & Sentiment & \cite{ahia2020towards}\\ 
  \multicolumn{1}{c}{} & \href{ https://zenodo.org/record/4300294#.YW8qHPrMJyx}{Swahili News Classification} & swa & Classification & \\ 
 
 \multicolumn{1}{c}{} & \href{ https://github.com/IsraelAbebe/An-Amharic-News-Text-classification-Dataset}{Amharic News classification} & amh & Classification & \cite{azime2021amharic}\\ 

  \multicolumn{1}{c}{} & \href{ https://github.com/uds-lsv/transfer-distant-transformer-african}{A study on African Language} & hau, yor & NER, TC & \cite{hedderich-etal-2020-transfer}\\ 
  
   \multicolumn{1}{c}{} & \href{ https://github.com/ajesujoba/YorubaTwi-Embedding}{ Y{o}r\`{u}b\'{a}Twi-Embedding} & aka, yor & NER, embedding & \cite{alabi2020massive}\\ 
   
  \multicolumn{1}{c}{} & & 40 languages including: \href{https://docs.google.com/uc?export=download&id=1RVRaSdwjuILTYFez-Nl73UMv2aubHzD6}{amh} \href{https://docs.google.com/uc?export=download&id=1lMkb_gYpwzd32_-waG_eNaWeehm0OpGZ}{hau} \href{https://docs.google.com/uc?export=download&id=1B5td0FABADD3xAEIWO_-ZwBuoV5kW85h}{ibo}, \\
  \multicolumn{1}{c}{} & \href{ https://github.com/csebuetnlp/xl-sum}{XL Sum} & \href{https://docs.google.com/uc?export=download&id=1DbPJGYoGAfclcvgWTgf0YP48u6gBdS-t}{kin}, \href{https://docs.google.com/uc?export=download&id=1ZCfG5L8A77P4BvOVm3O7BXQFliGyQCqY}{gaz}, \href{https://docs.google.com/uc?export=download&id=17n8UpuZSbWisvkPB7eklM3VARN3iNO6y}{pcm}, \href{https://docs.google.com/uc?export=download&id=1f9_4DjgTxmfhquJqnSe2x170FILsAiYi}{som}, \href{https://docs.google.com/uc?export=download&id=1WitOEsFdJdirZYZr92H0v0HGFpwi2vbh}{swa}, \href{https://docs.google.com/uc?export=download&id=1oHxPjk7PQ0JSkbf906wZYyK4pGdEulLL}{yor} & Summarization & \cite{hasan-etal-2021-xl}\\ 
  
  \multicolumn{1}{c}{} & & 282 languages including aar, afr, amh, \\
   \multicolumn{1}{c}{} &  & bam, ewe, Fula*, hau, ibo, \\
    \multicolumn{1}{c}{\multirow{-6}{*}{\rotatebox[origin=c]{90}{\textbf{\small Labelled}}}} & & kab, kon, kik, kua, kau, \\
   \multicolumn{1}{c}{} &  & lin, mlg, ndo, \\
   \multicolumn{1}{c}{} & & nso, gaz, run, kin, sna,\\
   \multicolumn{1}{c}{} &  & som, sot, ssw, swa, \\
   \multicolumn{1}{c}{} & \href{https://drive.google.com/drive/folders/1Q-xdT99SeaCghihGa7nRkcXGwRGUIsKN?usp=sharing/}{WikiAnn} & tsn, tso, wol, xho, yor, zul
   & NER &  \cite{pan2017cross}\\ 
 \multicolumn{1}{c}{} &  &  \href{https://catalog.ldc.upenn.edu/LDC2021T02}{aka}, \href{https://catalog.ldc.upenn.edu/LDC2018T04}{amh},
\href{https://catalog.ldc.upenn.edu/LDC2020T11}{gaz}, \\ 
\multicolumn{1}{c}{} &  DARPA LORELEI& \href{https://catalog.ldc.upenn.edu/LDC2018T11}{som},
\href{https://catalog.ldc.upenn.edu/LDC2020T22}{tir} & NER, SemAnal & \cite{strassel-tracey-2016-lorelei}\\ 
 \multicolumn{1}{c}{} & \href{https://github.com/Niger-Volta-LTI/yoruba-text}{Automatic Diacritic Restoration} & yor  & ADR &  \cite{orife2020improving}\\ \midrule
 
\multicolumn{1}{c}{} & &  10 languages including: afr, nya, hau, \\
\multicolumn{1}{c}{} & \href{https://www.tensorflow.org/datasets/catalog/c4#c4multilingual_nights_stay}{mC4} & ibo, sna, som, Sotho*, swa,xho, yor, zul & LM & \cite{xue-etal-2021-mt5} \\ 

\multicolumn{1}{c}{} & \href{https://zenodo.org/record/3553423#.YW9JlfrMJyx}{Swahili Language Modeling} &  swa & LM &  \\ 

\multicolumn{1}{c}{} && 1600+ (including 313 Niger-Congo), \\
\multicolumn{1}{c}{} & The John Hopkins University Bible Corpus &  67 Afro-Asiatic, and 52 Nilo-Saharan & LM & \cite{mccarthy-etal-2020-johns} \\

\multicolumn{1}{c}{} & \href{https://repo.sadilar.org/handle/20.500.12185/524}{Monolingual xho corpus} &  swa & LM & \\ 

\multicolumn{1}{c}{\multirow{-6}{*}{\rotatebox[origin=c]{90}{\textbf{\small Unlabelled}}}} & \href{https://oscar-corpus.com/}{OSCAR} & 166 languages including: \href{https://oscar-public.huma-num.fr/shuff-orig/af}{afr}, \href{https://oscar-public.huma-num.fr/shuff-orig/sw}{swa}, \href{https://oscar-public.huma-num.fr/shuff-orig/yo}{yor} & LM & \cite{suarez2020monolingual} \\\midrule

\multicolumn{1}{c}{} & \href{https://en.wikipedia.org/wiki/List_of_Wikipedias}{Wikipedia}& 37 African languages & LM & \\ 

\multicolumn{1}{c}{} & \href{https://alaroye.org/category/iroyin/}{Alaroye}& yor & LM & \\

\multicolumn{1}{c}{} & & \href{https://www.voaafaanoromoo.com/}{gaz}, \href{https://amharic.voanews.com/}{amh}, \href{https://www.voabambara.com/}{bam}, \href{https://www.voahausa.com/}{hau}, \href{https://www.radiyoyacuvoa.com}{kin}, 
\href{https://www.voalingala.com/}{lin}, \\ 
\multicolumn{1}{c}{} & \href{https://www.voanews.com/navigation/allsites}{VOA} &  \href{https://www.voandebele.com/}{nbl}, \href{https://www.voashona.com/}{sna}, \href{https://www.voasomali.com/}{som}, \href{https://www.voaswahili.com/}{swa}, \href{https://tigrigna.voanews.com/}{tir}, \href{https://www.voazimbabwe.com/}{Zimbabwe}& LM & \\ 

\multicolumn{1}{c}{} &\href{https://www.jw.org/en/}{Jehovah's witness} & More than $100$ African languages & MT& \cite{agic-vulic-2019-jw300}\\  

\multicolumn{1}{c}{} & &  \href{https://www.bbc.com/afaanoromoo}{gaz},  \href{https://www.bbc.com/amharic}{amh}, \href{https://www.bbc.com/hausa}{hau}, 
\href{https://www.bbc.com/igbo}{ibo}, \\
\multicolumn{1}{c}{\multirow{-6}{*}{\rotatebox[origin=c]{90}{\textbf{\small Raw Data}}}} &\href{https://www.bbc.co.uk/ws/languages}{BBC News} &\href{https://www.bbc.com/gahuza}{run},
\href{https://www.bbc.com/pidgin}{pcm},
 \href{https://www.bbc.com/somali}{som},
  \href{https://www.bbc.com/swahili}{swa},
   \href{https://www.bbc.com/tigrinya}{tir},
     \href{https://www.bbc.com/yoruba}{yor}
& LM & \\ \midrule

\multicolumn{1}{c}{} & \href{https://tanzil.net/trans/}{Tanzil} & \href{https://tanzil.net/trans/ber.mensur}{kab}, \href{https://tanzil.net/trans/am.sadiq}{amh}, \href{https://tanzil.net/trans/ha.gumi}{hau}, \href{https://tanzil.net/trans/so.abduh}{som}  \href{https://tanzil.net/trans/sw.barwani}{swa}& MT & \cite{TIEDEMANN12.463}  \\ 

\multicolumn{1}{c}{} & \href{https://zenodo.org/record/3734260#.YG4Kcc-Slyw}{Amharic Evaluation Dataset} & amh-eng & MT & \cite{hadgu2020evaluating}  \\ 

\multicolumn{1}{c}{} & \href{https://github.com/AAUThematic4LT/Parallel-Corpora-for-Ethiopian-Languages}{Parallel Corpora for Ethiopian Languages} & eng-amh, tir, gaz, wal, gez & MT & \cite{abate-etal-2018-parallel}\\ 

\multicolumn{1}{c}{} & \href{https://zenodo.org/record/4764039#.YW9pE_rMJyx}{English-Luganda Parallel Corpora} & eng-lug & MT & \\ 

\multicolumn{1}{c}{} & \href{https://gamayun.translatorswb.org/download/monosw-fr/}{Back-translated Swahili-French 1M sentence parallel data} & swa-fra & MT & \cite{oktem2021congolese} \\ 

\multicolumn{1}{c}{} & \href{https://gamayun.translatorswb.org/download/gamayun-5k-english-swahili/}{Gamayun Mini kit 5k} & swa-eng & MT &  \\ 

\multicolumn{1}{c}{} & \href{https://gamayun.translatorswb.org/download/gamayun-mini-kit-5k-kanuri-english/}{Gamayun Mini kit 5k} & kau-eng & MT &  \\ 

\multicolumn{1}{c}{} & \href{https://zenodo.org/record/4432117#.YW9rA_rMJyx}{English-Akuapem Twi parallel corpus} & eng-aka & MT &  \\ 

\multicolumn{1}{c}{} & & afr, amh, ful, hau, ibo, kea, \\
\multicolumn{1}{c}{} & & kam, luo, nso, nyj, gaz, som & MT &  \cite{Flores_2}\\
\multicolumn{1}{c}{} & \href{https://github.com/facebookresearch/flores?fbclid=IwAR1dkUPJ1XA4PibzNx9VM6wwRqUFXKV1Au1_NvDfGGEc4lYGX0pFKSY36N4}{FLORES-101} &   swa, wol, xho, yor, zul &  &  \cite{Flores_2}\\ 

\multicolumn{1}{c}{} & \href{https://opus.nlpl.eu/XhosaNavy.php}{Xhosa-English} & xho-eng & MT & \cite{TIEDEMANN12.463} \\  

\multicolumn{1}{c}{} & \href{https://catalog.ldc.upenn.edu/LDC2016L01}{Bamanankan Lexicon} & bam-eng & MT &\\  

\multicolumn{1}{c}{} & \href{https://repo.sadilar.org/handle/20.500.12185/404}{Autshumato} & eng-tsn & MT & \\ 

\multicolumn{1}{c}{} & & efi, afr, amh, bin, ddn, fon, \\
\multicolumn{1}{c}{} & & hau, ibo, ish, iso, kam, kik, kmb,\\ \multicolumn{1}{c}{} & & lin, lua, luo, nbl, nso, nya,  \\
\multicolumn{1}{c}{} & & pcm, sna, sot, swa, tir, tiv, tsn, \\
\multicolumn{1}{c}{} &\href{https://github.com/masakhane-io/masakhane-mt}{Masakhane MT} & aka, urh,  ven, xho, yor, zul, swc & MT & \cite{orife2020masakhane}\\ 

\multicolumn{1}{c}{\multirow{-6}{*}{\rotatebox[origin=c]{90}{\textbf{\small Parallel}}}} &\href{https://github.com/israaar/mt_bambara_data_models}{Bambara Dataset} & bam, eng and fra & MT & \cite{tapo2020neural} \\   
\multicolumn{1}{c}{} &\href{https://github.com/surafelml/Afro-NMT}{AFRONMT} & eng, swa, amh, tir, gaz, som & MT & \cite{lakew2020low}\\  
\multicolumn{1}{c}{} & \href{https://github.com/machelreid/afromt/blob/main/README.md}{AFROMT} &afr, xho, zul, run, sot, swa, bem, lin & MT & \cite{reid21afromt} \\ 

\multicolumn{1}{c}{} &  & 137 languages including \href{http://www.statmt.org/cc-aligned/en_XX-yo_NG.tsv.xz}{yor}, \href{http://www.statmt.org/cc-aligned/en_XX-af_ZA.tsv.xz}{afr}, \href{http://www.statmt.org/cc-aligned/en_XX-ak_GH.tsv.xz}{aka},   \href{http://www.statmt.org/cc-aligned/en_XX-am_ET.tsv.xz}{amh}, \\
\multicolumn{1}{c}{} & & \href{http://www.statmt.org/cc-aligned/en_XX-ff_NG.tsv.xz}{Fulfulde}, \href{http://www.statmt.org/cc-aligned/en_XX-ig_NG.tsv.xz}{ibo}, \href{http://www.statmt.org/cc-aligned/en_XX-so_SO.tsv.xz}{som}, \href{http://www.statmt.org/cc-aligned/en_XX-sw_KE.tsv.xz}{swa}, \\ \multicolumn{1}{c}{} & \href{http://www.statmt.org/cc-aligned/}{CCAligned} & \href{http://www.statmt.org/cc-aligned/en_XX-wo_SN.tsv.xz}{wol}, \href{http://www.statmt.org/cc-aligned/en_XX-yo_NG.tsv.xz}{yor}, \href{http://www.statmt.org/cc-aligned/en_XX-zu_ZA.tsv.xz}{zul} & MT & \cite{elkishky_ccaligned_2020} \\

\multicolumn{1}{c}{} & \href{https://github.com/IgnatiusEzeani/IGBONLP/tree/master/ig_en_mt}{IgboNLP} &  ibo-eng  & MT & \cite{ezeani2020igbo} \\

\multicolumn{1}{c}{} &\href{https://github.com/dadelani/menyo-20k_MT}{MENYO-20k} & yor-eng & MT & \cite{adelani2021menyo}  \\ 

\multicolumn{1}{c}{} &\href{https://www.findke.ovgu.de/findke/en/Research/Data+Sets/Amharic_English+Parallel+Corpus-p-1144.html}{Extended Amharic-English bilingual corpus}& amh, eng & MT & \cite{gezmu2021extended} \\

\multicolumn{1}{c}{} &\href{https://github.com/facebookresearch/LASER/tree/master/tasks/WikiMatrix}{WikiMatrix} & 85 languages including swa & MT & \cite{schwenk2019wikimatrix} \\

\multicolumn{1}{c}{} & Lorelei & \href{https://catalog.ldc.upenn.edu/LDC2021T02}{aka}& MT & \cite{schwenk2019wikimatrix} \\

\multicolumn{1}{c}{} &\href{https://paracrawl.eu/}{Paracrawl} & 39 languages including \href{https://s3.amazonaws.com/web-language-models/paracrawl/bonus/en-so.txt.gz}{som} and \href{https://s3.amazonaws.com/web-language-models/paracrawl/bonus/en-sw.txt.gz}{swa} & MT & \cite{banon-etal-2020-paracrawl} \\

\multicolumn{1}{c}{} & & 100 languages including afr, amh, \\
\multicolumn{1}{c}{} & & Coptic, din, ewe, kab, dop, som, swa, \\
\multicolumn{1}{c}{} &\href{http://christos-c.com/bible/}{Parallel Bible Corpus} &  shi, ttq, wal, wol, xho, xuu, zul & MT & C\&S \\

\multicolumn{1}{c}{} & \href{https://github.com/bonaventuredossou/ffr-v1}{FFR v1.1} & fon-fra & MT & \cite{dossou2020ffr} \\ \midrule

\multicolumn{1}{c}{} & \href{https://repo.sadilar.org/handle/20.500.12185/530}{SPCS Speech Corpus} &  eng, nso  & Speech & \cite{modipa2013implications} \\ 

\multicolumn{1}{c}{} & \href{https://repo.sadilar.org/handle/20.500.12185/530}{TTS data for four South African languages} &  afr, sot, tsn and xho & Speech & \\ 

\multicolumn{1}{c}{} & \href{https://github.com/besacier/mboshi-french-parallel-corpus}{Mboshi French Parallel Corpus} &  mdw, fra & Speech & \cite{DBLP:journals/corr/abs-1710-03501}\\

\multicolumn{1}{c}{} & \href{https://iwslt.org/2021/low-resource}{IWSLT Low Resource Shared Task} &  swh-eng, swc-fra & Speech & \cite{anastasopoulos-etal-2021-findings}\\ 

\multicolumn{1}{c}{} & \href{https://tico-19.github.io/}{Tico-19} &\href{https://gamayun.translatorswb.org/download/congolese-swahili-tico-19-audio-test-set/}{swc} & Speech & \cite{anastasopoulos-etal-2020-tico}\\ 
     \multicolumn{1}{c}{} & GlobalPhone &  \href{http://catalog.elra.info/en-us/repository/browse/ELRA-S0347/}{hau}, \href{http://catalog.elra.info/en-us/repository/browse/ELRA-S0375/}{Swahili}  & Speech & \cite{schultz2002globalphone} \\ 
     
   \multicolumn{1}{c}{} & The NCHLT Speech Corpus & \href{https://repo.sadilar.org/handle/20.500.12185/280}{afr}, \href{https://repo.sadilar.org/handle/20.500.12185/277}{tso}, \href{https://repo.sadilar.org/handle/20.500.12185/281}{tsn}, \href{https://repo.sadilar.org/handle/20.500.12185/278}{sot}, \href{https://repo.sadilar.org/handle/20.500.12185/270}{nso}, \\
   \multicolumn{1}{c}{} & of the South African languages & \href{https://repo.sadilar.org/handle/20.500.12185/275}{zul}, \href{https://repo.sadilar.org/handle/20.500.12185/276}{ven}, 
   \href{https://repo.sadilar.org/handle/20.500.12185/271}{ssw}, 
   \href{https://repo.sadilar.org/handle/20.500.12185/279}{xho}, \href{https://repo.sadilar.org/handle/20.500.12185/272}{nbl} & Speech & \cite{NCHLT}\\ 
    \multicolumn{1}{c}{} & \href{http://alffa.imag.fr/}{ALFFA} &  amh, swh, hau, wol & Speech & \cite{gauthier-etal-2016-collecting} \\ 
     \multicolumn{1}{c}{} & \href{https://github.com/laleye/ALFFA_PUBLIC/tree/master/ASR/FONGBE}{Fon ASR} & fon & Speech & \cite{laleye:hal-01436788} \\ 
     \multicolumn{1}{c}{\multirow{-6}{*}{\rotatebox[origin=c]{90}{\textbf{\small Speech}}}} & \href{https://gamayun.translatorswb.org/download/swahili-audio-mini-kit/}{Swahili audio mini-kit} & swh & Speech &  \\ 
     \multicolumn{1}{c}{} & \href{https://coqui.ai/swahili-congo/twb/v0.3.0}{Swahili (Congo) STT v0.3.0 } & swc & Speech &  \cite{swc-stt}\\ 
    \multicolumn{1}{c}{} & \href{https://aimsammi.org/}{AIMS} &  hau, lug, kab, kin & Speech & \cite{ DBLP:journals/corr/abs-2103-08993} \\ 
 \multicolumn{1}{c}{}  & & \href{https://catalog.ldc.upenn.edu/LDC2019S22}{amh}, \href{https://catalog.ldc.upenn.edu/LDC2019S16}{ibo}, \href{https://catalog.ldc.upenn.edu/LDC2020S02}{luo}, \\
 \multicolumn{1}{c}{} &  IARPA Corpus & \href{https://catalog.ldc.upenn.edu/LDC2019S22}{amh}, \href{https://catalog.ldc.upenn.edu/LDC2019S16}{ibo}, \href{https://catalog.ldc.upenn.edu/LDC2020S02}{luo}, \href{https://catalog.ldc.upenn.edu/LDC2017S05}{swh}, \href{https://catalog.ldc.upenn.edu/LDC2017S19}{zul} & Speech & \cite{inproceedings} \\ \bottomrule

\end{tabular}}
\end{adjustbox}
\caption{List of available data resources. TC=Topic Classification. C\&S=\cite{christodouloupoulos2015massively}.}\label{tab:resources}
\end{table*}



\begin{table*}[h!] 
\footnotesize

\centering
\begin{adjustbox}{max width=\textwidth}
\renewcommand{\arraystretch}{1.0}
\begin{tabular}{p{2.5cm} p{2.5cm} p{2.5cm} p{2.5cm} p{2.5cm} p{2.5cm}p{2.5cm}} 
\toprule
\thead{\textbf{Lang }}  & \thead{\textbf{Lang }} &  \thead{\textbf{Lang }} &  \thead{\textbf{Lang }}  & \thead{\textbf{Lang }}  & \thead{\textbf{Lang }} & \thead{\textbf{Lang }} \\ \midrule
K\'{a}sim,0 & isekiri , 0 & ndonga , 0 & matuumbi, 0 & b\'{e}t\'{e} , 0 & bini,   0  &babole ,   0 \\ 
obolo,   0 & ghulfan,   0 & masakin ,   0 & alagwa ,   0 & tem ,   0 & miisiirii , 0  & gokana ,   0\\
baga sitemu ,   0 & vagla ,   0 & mundani ,   0 & mbole ,   0 & kom ,   0 & ndut,   0   & gurenne ,   0 \\
hemba,   0 & gbeya bossangoa ,   0 & seychelles creole ,   0 & grebo ,   0 & guere ,   0& majang , 0 & waama ,   0 \\  
bujeba ,   0 & ewondo ,   0 & mankanya ,   0 & emai ,   0 & moro ,   0 & lamé ,   0 &shatt ,   0 \\
kohumono ,   0 & tetela ,   0 & baka ,   0 & qafar ,   0 & wan ,   0 &talinga ,   0& soninke ,   0 \\
gbaya kara ,   0 & yaka ,   0 & bororo ,   0 & vili ,   0 & tennet ,   0  & palor ,   0 &buduma ,   0 \\
balanta ,   0 & bai ,   0 & mandinka ,   0 & mango ,   0  & iraqw ,   0& ajagbe ,   0  & bafut ,   0 \\
nubi ,   0 & migama ,   0 & burunge,   0 & bobo madaré ,   0 & lobi ,   0 &  yamba ,   0  & tera ,   0 \\
manjaku ,   0 & tommo so ,   0 & otoro ,   0 & shuri ,   0 & dyula ,   0 & tenyer ,   0   & koyraboro senni ,   0 \\
comorian ,   0 & duma ,   0 & mamvu ,   0 & hamer ,   0 & kasem ,   0 & mara ,   0  & temne ,   0\\
bankon ,   0 & kisi ,   0 & sama ,   0 & yeyi ,   0  & tuki ,   0  & kxoe ,   0   & guduf ,   0\\

kwangali ,   0& supyire ,   0 & dangaléat ,   0 & mofu-gudur ,   0  &mokilko ,   0 & tigré ,   0 & ful ,   0 \\
bandi ,   0 & herero ,   0 & !xun ,   0 & bangime ,   0 & tuareg ,   0 & mbe' ,   0 & mayogo ,   0 \\
ko ,   0 & sena ,   0 & chumburung ,   0 & bafia ,   0 & bori ,   0  & kilba ,   0  & avokaya ,   0 \\
ejagham ,   0 & londo ,   0 & avatime ,   0 & sisaala ,   0 & ma'di ,   0 & bakundu ,   0  & nyimang ,   0 \\
darma ,   0 & tunen ,   0 & wolaytta ,   0 & mbodomo ,   0 & mupun ,   0  & kenyan S.L ,   0 &gamo ,   0 \\
ciluba ,   0 & turkana ,   0 & sungor ,   0 & uma ,   0 & degema ,   0 & akwa ,   0& aghem ,   0  \\
kpelle ,   0 & päri ,   0 & tamashek ,   0 & aizi ,   0 &katcha ,   0 & ijo , 0 & baale ,   0\\
hunde ,   0 & samba leko ,   0 & ngizim ,   0 & príncipense ,   0 & nupe ,   0 & tumak ,   0  & nc\`{a}m ,   0 \\
me'en ,   0& duala ,   0 & ghotuo ,   0 & ik ,   0 &mwera ,   0 &  kanakuru ,   0 & nsenga ,   0 \\
kera ,   0 & seme ,   0 & bidiya ,   0 & birri ,   0  & fongbe ,   0 & jukun ,   0 & burum ,   0 \\
bobangi ,   0 & ekoti ,   0 & midob ,   0 & mbugu ,   0  & aja ,   0 & sukumam ,   0 &tama ,   0\\
hadza ,   0 & ugandan S.L ,   0 & bushoong ,   0 & maninka ,   0&efik ,   0  & kotoko ,   0  &kukú ,   0  \\
kunama ,   0 &  rundi ,   0 & muher ,   0 & mauka ,   0 & lua ,   0 & moru ,   0  & avikam ,   0 \\
daba ,   0 & mundang ,   0 & dongo ,   0 & beembe ,   0 & mankon ,   0 & toro so ,   0 & krongo ,   0\\
bamun ,   0 & tiv ,   0 & wobe ,   0 & miya ,   0 & diola-fogny ,   0 & mbili ,   0  & basa\'{a} ,   0\\
kuanyama ,   0 & sebei ,   0 & karimojong ,   0 & orig ,   0  & budu ,   0 & sandawe ,   0  & yakoma ,   0  \\
laal ,   0 & pero ,   0 & //ani ,   0 & awngi ,   0 & kete ,   0  & daju ,   0 & lebeo ,   0  \\
leko ,   0 & mambwe ,   0 & lango ,   0 & hdi ,   0 &shinassha ,   0 & songe ,   0  & mpongwe ,   0\\
bimoba ,   0 & ogbronuagum ,   0 & bayso ,   0 & kinga ,   0 & acholi ,   0 &bilin ,   0 & chaga ,   0 \\
nara ,   0 & dizi ,   0 & nyanga,   0 & jeli ,   0 & hehe ,   0& pokot ,   0 & burji ,   0\\
enya ,   0 & mano ,   0 & nharo ,   0 & baule ,   0 & maasai ,   0 & mondunga ,   0  & kagoma ,   0\\
ngbandi ,   0 & lendu ,   0 & tirmaga ,   0 & leti ,   0  & nande ,   0& runyankore ,   0  & shambala ,   0  \\
fyem ,   0 & yemsa ,   0 & lafofa ,   0 & ingessana ,   0  & nandi ,   0& lele ,   0 &  senadi ,   0  \\
mituku ,   0 & gula iro ,   0 & fur ,   0 & kirma ,   0 & fe'fe' ,   0 &gula ,   0 & niuafo'ou ,   0 \\
malgwa ,   0 & ebira ,   0 & berber ,   0 & ju|'hoan ,   0  & mono ,   0 & ama ,   0 &ngambay ,   0  \\
bura-pabir ,   0 & gusii ,   0 & bolia ,   0 & buli ,   0 &sangu ,   0 & ika ,   0 & shabo ,   0\\
kele ,   0 & kullo ,   0 & nkem ,   0 & gan ,   0 & beria ,   0  &nkonya ,   0  & langi ,   0 \\
izi ,   0 & makonde ,   0 & bariba ,   0 & babungo ,   0 & kposo ,   0 & giziga ,   0 & oku ,   0  \\
mongo ,   0 & !xóõ ,   0 & jomang ,   0 & kenga ,   0 & temein ,   0  & Kami ,   0 & gorowa ,   0 \\ 
ding ,   0 & kalanga ,   0 & coptic ,   0 & urhobo ,   0 & gumuz ,   0 & gunu ,   0 & bukusu ,   0 \\
dagaare ,   0 & uldeme ,   0 & gworok ,   0 & afar ,   0 & bakueri ,   0  & bana ,   0 & karó ,   0 \\
tampulma ,   0 & mende ,   0 & lunda ,   0 & haya ,   0 & nkore-kiga ,   0& guinea bissau c. ,   0 & amele ,   0 \\
neyo ,   0 & bira ,   0 & fulfulde ,   0 & kanyok ,   0  & bahnar ,   0  & miri ,   0& nyiha ,   0 \\
bodo ,   0 & lelemi ,   0 & logoti ,   0 &  mbalanhu ,   0 & munzombo ,   0 & kenyang ,   0 & dabida ,   0 \\
bozo,   0 & karanga ,   0 & bisa ,   0 & konyagi ,   0& tashlhiyt ,   0  & ndebele ,   0 & dullay ,   0  \\
mbosi ,   0 & goemai ,   0 & murle ,   0 & =|hoan ,   0 & teso ,   0 & ngbaka ,   0 & kefa ,   0  \\
ndogo ,   0 & ronga ,   0 & tonga ,   0 & kresh ,   0 & gungbe ,   0 & bubi ,   0 & koranko ,   0 \\
konni ,   0 & guro ,   0 & mambila ,   0 & mündü ,   0 & da'a ,   0& nuer ,   0 & runyoro-rutooro ,   0  \\
maale ,   0 & dhaasanac ,   0 & angas ,   0 & harari ,   0 & bagiro ,   0& bade ,   0 & ngoni ,   0 \\
ibibio ,   0 & pa'a ,   0 & mooré ,   0 & lozi ,   0  & toussian ,   0& nzakara ,   0  & rimi ,   0 \\
zayse ,   0 & gimira ,   0 & birom ,   0 & leggbó ,   0 & benga ,   0 &lagwan ,   0 & margi ,   0  \\ 
pangwa ,   0 & zande ,   0 & isoko ,   0 & mampruli ,   0& kpan ,   0 & masalit ,   0 & konkomba ,   0\\
gola ,   0 & beng ,   0 & maba ,   0 & nyangi ,   0 & ngemba ,   0 & saho, 0 & suku ,   0 \\
musgu ,   0 & adioukrou ,   0 & /xam ,   0 & tikar ,   0 & broken ,   0 & yana ,   0 & mada ,   0\\
nuni ,   0 & binga ,   0 & kagulu ,   0 & ndumu ,   0& holoholo ,   0 &jur mödö ,   0  & mumuye ,   0 \\

nyamwezi ,   0 & shilluk ,   0 & ron ,   0 & dime ,   0 & ngombe ,   0 & buma ,   0 & dahalo ,   0 \\
dhivehi ,   0 & kosop ,   0 & defaka ,   0 & bongo ,   0 & luwo ,   0  & lugbara ,   0 & koyra chiini ,   0 \\
kituba ,   0 & dii ,   0 & abidji ,   0 & boko ,   0 & komo ,   0  & lamnso' ,   0 & klao ,   0 \\
kadugli ,   0 & kabiyé ,   0 & nyambo ,   0 & mbum ,   0 & bole ,   0& linda ,   0 &ila ,   0 \\
ntomba ,   0 & lese ,   0 & luvale ,   0 & lyele ,   0 & busa ,   0 & doko ,   0 & igede ,   0 \\
aka ,   0 & nateni ,   0 & idoma ,   0 & kara ,   0 & n'ko ,   0  & khoekhoe ,   0 & rendille ,   0 \\
katla ,   0 & tabwa ,   0 & korana ,   0 & koh ,   0 & pogoro ,   0 & didinga ,   0& luri ,   0  \\
vata ,   0 & podoko ,   0 & yulu ,   0 & tangale ,   0 & lamang ,   0& engenni ,   0 & dadjriwalé ,   0   \\
berta ,   0 & tsogo ,   0 & dagbani ,   0 & bulu ,   0 &kiluba ,   0 & tarok ,   0 & datooga ,   0  \\
bari ,   0 & mungaka ,   0 & ega ,   0 & ifumu ,   0 & mahican ,   0 & gude ,   0&  runga ,   0 \\
sare ,   0 & masa ,   0 & yansi ,   0 & mbay ,   0 & wéménugbé ,   0  & sengele ,   0 & kela ,   0\\
anyi ,   0 & fulani ,   0 & mursi ,   0 & soddo ,   0 & diola-kasa ,   0 &jamsay ,   0 & koorete ,   0 \\
ga'anda ,   0 & arbore ,   0 & anywa ,   0 & loma ,   0 & fiote ,   0 & dyimini ,   0    & alladian ,   0\\
bena-lulua ,   0 & mbere ,   0 & doyayo ,   0 & gidar ,   0 & etsako ,   0 & ngiti ,   0  & ogbia ,   0 \\
subiya ,   0 & mba ,   0 & chai ,   0 & tupuri ,   0 & kanembu ,   0 & tima ,   0 & koromfe ,   0 \\
godié ,   0 & nanerge ,   0 & mambai ,   0 & koegu ,   0 & lucazi ,   0 &adamorobe S.L ,   0 & anufo ,   0\\
sotho ,   0 & dong ,   0 & aari ,   0 & kemant ,   0 & kanuri ,   0 & sidaama ,   0  & donno so ,   0  \\
bemba ,   0 & deti ,   0 & lamba ,   0 & angolar ,   0 & gade ,   0& gunya ,   0  & barambu ,   0\\
bagirmi ,   0 & kamba ,   0 & mbara ,   0 & vai ,   0 & makaa ,   0 & gwari ,   0  & nafaanra ,   0 \\
nigerian pidgin ,   0  & rashad ,   0 & mangbetu ,   0 & somali , 1 & igbo , 1 & bambara, 1 & venda , 1\\
 tumbuka , 1 &twi, 1 & sango, 1 & kikuyu, 1 & kirundi, 1 & ndonga, 1 & lingala, 1\\
 sesotho , 1 & chichewa , 1 & dinka , 1  & malagasy , 1 & ewe , 1 & kinyarwanda , 1 & kabiye , 1 \\ 
kongo , 1 & northern sotho , 1 & kabyle , 1 & oromo , 1 & akan , 1 & tsonga , 1 & luganda , 1 \\
amharic , 2 & hausa , 2 & xhosa , 2 & swahili , 2 & zulu , 2 & tswana , 2 & wolof , 2 \\
tigrinya , 2 & Y{o}r\`{u}b\'{a} , 2 & afrikaans , 3 \\ \bottomrule


 \end{tabular}
 \end{adjustbox}
 \caption{\footnotesize{Language diversity index. Adapted from \newcite{joshi-etal-2020-state}}.} \label{tab:l_index}
 \end{table*}

\section{Data Quality}\label{sec:dataquality}
The preliminary evaluation of Flores101 dataset for Y{o}r\`{u}b\'{a} was done by a native speaker of Y{o}r\`{u}b\'{a} who is also a linguist. Specifically, $57\%$ of the dataset was randomly selected while keeping track of the word's sentential context and the English source context. We removed all numerals written with digits from the dataset before the random selection. This was to help us focus on lexical items alone. We found \textbf{(1)} spelling errors, \textbf{(2)} inconsistent spellings, which are instances of different spellings for the same word within the text, \textbf{(3)} borrowed words not adapted to the orthographic convention of the target language, without recourse to named entities, and \textbf{(4)} incorrect tone marks. Further evaluation will be required to access the quality of the dataset on a semantic and syntactic level. Examples of each of the errors identified is presented in Table \ref{tab:error-analysis}.
\begin{table*}[h!]
\small
\centering
\begin{adjustbox}{max width=\textwidth}
\renewcommand{\arraystretch}{1.15}
\begin{tabular}{p{3cm} p{11cm}}
\hline
 \thead{\textbf{Output}} & \thead{\textbf{Sentence}} \\
 \hline \hline
 \thead{\textbf{Spelling errors}} \\
 \hline 
 Y{o}r\`{u}b\'{a} Source & "Mo \textbf{d\'{u}p\'{e}} l\textsubdot{\'{o}}w\textsubdot{\'{o}} \`{a}d\textsubdot{o}  t\'{o} gb\'{o}r\`{u}k\`{u} ti \textsubdot{e}l\textsubdot{\'{e}}w\textsubdot{\`{o}}n b\'{i}i t\'{e}mi" ... \\ 
English Source & "Thanks for those who supported a convict like me",... \\ 
Y{o}r\`{u}b\'{a} Target & \textbf{\`{I}flr\'{u}n\'{u}} h\`{a}n b\textsubdot{\`{e}}r\textsubdot{\`{e}} n\'{i} ago m\textsubdot{\'{o}}k\`{a}nl\'{a} (UTC+1) n\'{i} Whitehall n\'{i} w\'{a}j\'{u} \textsubdot{e}nu \textsubdot{\`{o}}n\`{a} il\'{e}\ i\textsubdot{s}\textsubdot{\'{e}} \textsubdot{o}l\textsubdot{\'{o}}p\`{a}\'{a} s\'{i} \`{o}p\'{o}p\'{o}n\`{a} Downing, il\'{e} \'{a}r\textsubdot{e} or\'{i}l\textsubdot{\`{e}} `{e}d`{e}.\\
 English Source & The protest started around 11:00 local time (UTC+1) on Whitehall opposite the police-guarded entrance to Downing Street, the Prime Minister's official residence. \\ \hline \hline
  \thead{\textbf{Inconsistent spellings}} \\
 \hline
 Y{o}r\`{u}b\'{a} Target & Fidali, omo odun-28 ti darap\textsubdot{\`{o}} m\textsubdot{\'{o}} \textsubdot{e}gb\textsubdot{\'{e}} \textbf{agb\'{a}boolu} Basilona ... \\ 
 English Source & 28-year-old Vidal had joined Barça ...
 \\ 

  Y{o}r\`{u}b\'{a} Target & \textbf{Agb\'{a}b\textsubdot{\`{o}}l\`{u}} T\`{o}n\'{i} n\'{i} Alex Overchkin ti Washington Capitals. \\ 
 English Source & Today's Player of the Day is Alex Ovechkin of the Washington Capitals. \\
 
   Y{o}r\`{u}b\'{a} Target & K\'{o}s\textsubdot{\'{e}}l\`{o}m\'{i}n t\'{o} \textsubdot{s}er\'{e} j\`{u} t\`{a}b\'{i} j\textsubdot{e} g\'{o}\`{o}l\`{u} j\`{u} f\'{u}n ik\textsubdot{\`{o}} \textbf{Agb\'{a}b\textsubdot{\`{o}}\textsubdot{\`{o}}l\`{u}} ju Bobek \\ 
 English Source& No one else has ever made more appearances or scored more goals for the club than Bobek.
  \\  \hline \hline
 \thead{\textbf{Borrowed words not adapted to orthographic conventions of target language}} \\
 \hline
 Y{o}r\`{u}b\'{a} Target & \`{A}w\textsubdot{o}n kan gb\`{a}gb\textsubdot{\'{o}} p\textsubdot{\`{e}}l\'{u} john Grant, p\'{e} \`{a}ti \textbf{funding crunch} \`{a}ti s\'{i}s\'{u}n n\'{i} \textsubdot{\`{e}}k\textsubdot{\'{o}} \`{i}m\`{o}ye \`{e}t\'{o} or\'{i} am\'{o}h\`{u}nm\'{a}w\`{o}r\'{a}n d\'{a}si l\'{a}ti par\'{i} er\'{e} n\'{a}\`{a}. \\
 English Source & It is believed by some, including John Grant, that both the funding crunch and a shift in the philosophy of educational television programming contributed to ending the series. \\
 
 Yoruba Target & \`{A}w\textsubdot{o}n  on\'{i}m\textsubdot{\`{o}} s\'{a}y\textsubdot{\'{e}}ns\textsubdot{\`{i}} ma n p\`{e} n\'{i} \textbf{“stimulated emission of radiation"} tor\'{i} \`{a}w\textsubdot{o}n \'{a}t\textsubdot{\'{o}}m\textsubdot{s}ok{\`{u}} ma \'{n} fura s\'{i} in\'{a} t\'{o} r\`{a}n \`{e}y\'{i} \textsubdot{s}ok\`{u}n fa k\'{i} fotoni ina maa j\'{a}de, in\'{a} d\textsubdot{\`{e}} j\textsubdot{\'{e}} ir\'{u}f\textsubdot{\'{e}} redie\textsubdot{s}\textsubdot{\'{o}}ni. \\
 English Source & Scientists call this process "stimulated emission of radiation" because the atoms are stimulated by the bright light, causing the emission of a photon of light, and light is a type of radiation.
\\ 
 \hline \hline
  \thead{\textbf{Incorrect tone marking}} \\
 \hline
  Y{o}r\`{u}b\'{a} Target & \textbf{Awon iIwe naa fihan ile ifowopamo merinla to ran awon onisowo olola pa ilopo bilioni owo Amerika mo lati le sa fun owo ori ati awon ofin miin.} \\ 
 English Source & The documents showed fourteen banks helped wealthy clients hide billions of US dollars of wealth to avoid taxes and other regulations.
\\  \hline \hline 
 

\end{tabular}

\caption{Some errors from flores101 for Y{o}r\`{u}b\'{a}. We indicate the errors with bold type fonts. }\label{tab:error-analysis}
\end{adjustbox}
\end{table*}
\end{document}